\documentclass{bmvc2k}

\usepackage{bm}
\usepackage{amsfonts}
\usepackage{amssymb}
\usepackage{graphicx}
\usepackage{graphicx}
\usepackage{hyperref}
\usepackage{algorithm}
\usepackage{algpseudocode}
\usepackage{booktabs}
\usepackage{multirow}
\usepackage{threeparttable}
\usepackage[table,xcdraw]{xcolor}
\usepackage{pifont}
\usepackage{verbatim}
\usepackage{cleveref}

\newcommand{\cmark}{\ding{51}}%
\newcommand{\xmark}{\ding{55}}%
\newcommand{\argmax}{\operatornamewithlimits{argmax}}

\algnewcommand\algorithmicinput{\textbf{Input:}}
\algnewcommand\INPUT{\item[\algorithmicinput]}
\algnewcommand\algorithmicoutput{\textbf{Output:}}
\algnewcommand\OUTPUT{\item[\algorithmicoutput]}

\title{Discovering Class-Specific Pixels for Weakly-Supervised Semantic Segmentation}

\addauthor{Arslan Chaudhry}{arslan.chaudhry@eng.ox.ac.uk}{1}
\addauthor{Puneet K. Dokania}{puneet@robots.ox.ac.uk}{1}
\addauthor{Philip H.S. Torr}{philip.torr@eng.ox.ac.uk}{1}

\addinstitution{
 Department of Engineering Science\\
 University of Oxford\\
 United Kingdom
}

\runninghead{Chaudhry et al.}{DCSP}

\begin{document}

\maketitle

\begin{abstract}
We propose an approach to discover class-specific pixels for the weakly-supervised semantic segmentation task. We show that properly combining saliency and attention maps allows us to obtain reliable cues capable of significantly boosting the performance. First, we propose a simple yet powerful hierarchical approach to discover the class-agnostic salient regions, obtained using a salient object detector, which otherwise would be ignored. Second, we use fully convolutional attention maps to reliably localize the class-specific regions in a given image. We combine these two cues to discover class-specific pixels which are then used as an approximate ground truth for training a CNN. While solving the weakly supervised semantic segmentation task, we ensure that the image-level classification task is also solved in order to enforce the CNN to assign at least one pixel to each object present in the image. Experimentally, on the PASCAL VOC12 val and test sets, we obtain the mIoU of $60.8\%$ and $61.9\%$, achieving the performance gains of $5.1\%$ and $5.2\%$ compared to the published state-of-the-art results. The code is made publicly available.
\end{abstract}

\vspace{-5mm}
\section{Introduction}
\vspace{-3mm}
\label{sec:intro}
Convolutional Neural Networks (CNNs) are extremely successful in solving structured output prediction tasks such as semantic segmentation \cite{chen2014semantic, chandra2016fast, long2015fully, zheng2015conditional}, where the goal is to assign a semantic class label to each pixel. The prediction accuracy of CNNs in these tasks is heavily reliant on the large amounts of pixel-level annotated datasets \cite{everingham2010pascal, lin2014microsoft}. The collection of such datasets is an extremely laborious task -- it takes almost four minutes on average to annotate all the pixels in an image~\cite{everingham2010pascal, bearman2016s}. Additionally, pixel-level annotation becomes an impediment when it comes to scaling the segmentation networks to new object categories. 

To counter this curse of pixel-level annotation, recently, the focus has shifted towards weakly- and semi-supervised semantic segmentation methods which require reduced level of annotations. These methods incorporate any one or more of the following supervisions: image labels, bounding boxes, squiggles, spots etc.~\cite{HouMining2016, wei2017object, pinheiro2015image, wei2016stc, qi2016augmented, papandreou2015weakly, pathak2015constrained, kolesnikov2016seed}. Among these supervisions {\em image-level labels} are easiest to collect -- almost $1$ second per class or object-category ~\cite{papadopoulos2014training} -- and also are amenable to webly-supervised learning where one can download millions of images of new object categories from the Internet for training. Hence, in this work we focus on the image level labels-based supervision. 

Concretely, we combine so called attention and saliency cues to discover class-specific pixels in images that act as approximate/ weak ground-truth for training. Here the term {\em attention} is used to refer to the pixels in an image change in which affects the score of the class to be classified the most. There are different ways to localize these kind of discriminant pixels in an image. Motivated by~\cite{zhou2016CAM} we use global average pooling based classifier architecture to localize the discriminant pixels. We extend~\cite{zhou2016CAM} to a fully convolutional setting to get multi-object dense attention maps. We call this network a Fully Convolutional Attention Network (FCAN) (\cref{sec:fcan}). Note that the FCAN is trained using only image labels, and the attention maps we obtain are class-specific. 

We use the term {\em Saliency} to refer to the binary masks that detect visually noticeable foreground objects in an image. These masks are class-agnostic and provide complimentary information to the class-specific attention maps as they focus only on the foreground objects. In particular we use a salient object detector~\cite{liu2016dhsnet} to obtain salient region masks. One major limitation of such salient object detectors is their inability to detect multiple salient objects in an image. We propose an Hierarchical Saliency method (\cref{sec:hierSal}) that employs an iterative erasing strategy to rectify this problem. The saliency detector~\cite{liu2016dhsnet} is trained using class-agnostic salient region masks. 

The attention cues, obtained from the FCAN, focus only on the most discriminative part of an object, and do not provide any information on the extent of the object. On the other hand, the saliency cues give objectness information but are class-agnostic. We combine attention and saliency maps to obtain pixel-level class-specific approximate ground-truth to train a segmentation network.

Our training objective consists of a segmentation loss and an auxiliary classification loss. As the training progresses, we adapt (update) the pixel-level cues. 
The intuition behind the adaptive approach is that as the network trains under a finer loss function (pixel-wise cross-entropy), the localization cues must improve (experimentally verified) and, hence, it makes sense to iteratively update them. Given that the saliency maps can be obtained using any off-the-shelf saliency detector, our approach is end-to-end trainable.   

With this very simple technique, we obtain the mIoU of $60.8\%$ and $61.9\%$ on the PASCAL VOC 2012 $val$ and $test$ sets for the weakly supervised semantic segmentation task using image labels, achieving new state-of-the-art results. 

\vspace{-5mm}
\section{Related Works}
\vspace{-3mm}
Papandreou et al.~\cite{papandreou2015weakly} employed Expectation-Maximization to solve weakly-supervised semantic segmentation using annotated bounding boxes and image labels. Similarly, Hou et al.~\cite{HouMining2016} also relied on an EM inspired approach, however, they used image labels and saliency masks for the supervision. Di et al.~\cite{lin2016scribblesup} make use of scribbles to train the segmentation network where scribbles provide few pixels for which the ground truth labels are known. 
Similarly, Bearman et al.~\cite{bearman2016s} combines annotated points with objectness priors as the supervisory signals. Some approaches employ only image labels such as Pathak et al.~\cite{pathak2015constrained} and Pinheiro et al.~\cite{pinheiro2015image}. Pathak et al. framed the segmentation problem as a constrained optimization problem, whereas, Pinheiro et al. posed the problem as a multiple instance learning problem. Wei et al.~\cite{wei2016stc} proposed  a simple to complex framework where a network is first trained using simple images (single object category) followed by training over complex ones (multiple objects). Qi et al.~\cite{qi2016augmented} proposed to link semantic segmentation and object localization with proposal selection module, where generated proposals came from MCG~\cite{arbelaez2014multiscale}.  
Kolesnikov and Lampert~\cite{kolesnikov2016seed} proposed multiple loss functions that can be combined to improve the training. 
Recently, Wei et al.~\cite{wei2017object} proposed an adversarial erasing scheme in order to obtain better attention maps which in turn provide better cues for the training. 

Our work is closest to~\cite{HouMining2016,wei2017object}, but in contrast to \cite{wei2017object}, we do not employ erasing to expand attention maps which requires retraining of an attention/classification network after each erasing. Instead, we erase to discover new salient regions and keep the attention network intact. This way, the same saliency network can be used after each erasing to discover new salient regions. 
Additionally, instead of using different networks for the attention and segmentation tasks, as done by~\cite{wei2017object}, we use a single network and train it end-to-end for both the tasks. This helps us in progressively obtaining better attention cues. Similar to~\cite{HouMining2016} we employ attention and saliency based cues. However, ~\cite{HouMining2016} considered a simpler case -- images with a single object category -- and did not extend these cues for images with multiple objects.

\begin{figure}[]
\centering
\includegraphics[scale=0.13]{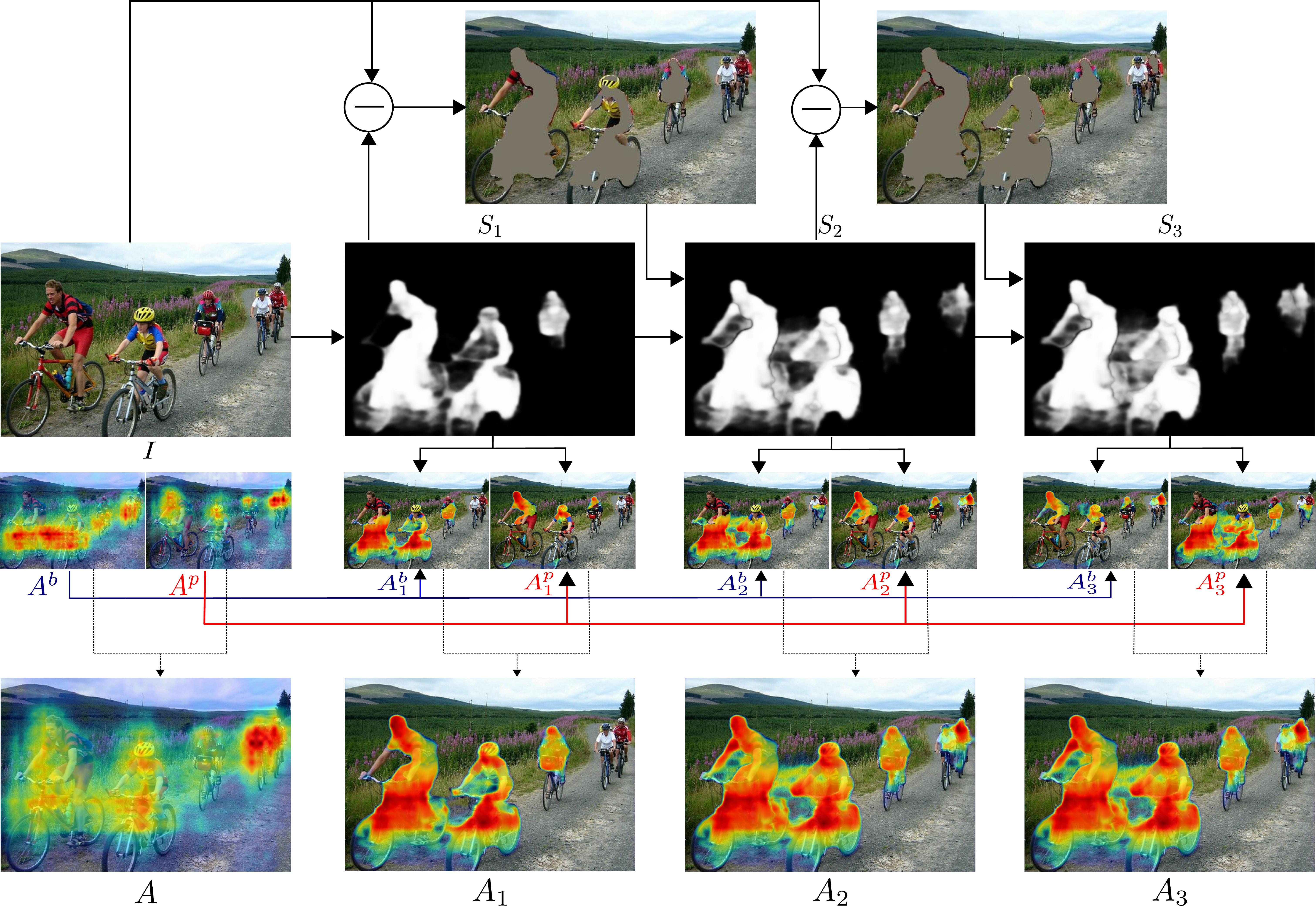}
\caption{Discovering Class-Specific Pixels: $I$, $A$, and $S_1$ represent the input image, the fully convolutional attention map (for both `bike' and `person', see Section~\ref{sec:fcan}) and the initial saliency map~\cite{liu2016dhsnet}. $S_2$ and $S_3$ represent the saliency maps obtained after first and second erasing. Superscripts `b' and `p' are used to show `bike' and `person' specific cues. Notice that, in the case of $S_2$ and $S_3$, more salient objects are discovered (for example, objects in the top right of the image). $A_1$, $A_2$ and $A_3$ represents the attention maps obtained using the combination of $A$ with $S_1$, $S_2$ and $S_3$, respectively. Comparing $A$ and $A_3$, it is evident that the attention map has improved significantly. Also, many false activations are removed and class-specific pixels are being discovered with high confidence.}
\label{fig:hierSal}
\end{figure}

\vspace{-6mm}
\section{Preliminaries}
\label{sec:prelim}
\vspace{-3mm}
\paragraph{\textcolor{bmv@sectioncolor}{Saliency}} There exists multiple definitions of saliency in the computer vision literature. The eye-fixation view~\cite{li2014secrets} of saliency computes a probabilistic map of an image to predict actual human eye gaze patterns. Alternatively, the salient object detection view generates a binary mask that detects important regions from natural images~\cite{shi2016hierarchical}. In this work, we employ the latter definition of saliency (see the second row in Figure~\ref{fig:hierSal}) and explicitly use~\cite{liu2016dhsnet} as our baseline saliency detector.  
\vspace{-5mm}  
\paragraph{\textcolor{bmv@sectioncolor}{Attention Map}} Similar to saliency, attention is also a vaguely defined term in the literature. The definition that we use treats attention as a set of pixels in an image towards which the CNN is most sensitive while classifying the image belonging to a certain object category. Formally, given an image $I$ consisting of $m$ object categories, the attention map ($A^c$) assigns a score $\in [0,1]$ to each pixel representing the likeliness of the pixel belonging to the $c$-th object category (see the third row in Figure~\ref{fig:hierSal}). 
\vspace{-4mm}
\paragraph{\textcolor{bmv@sectioncolor}{Weakly-Supervised Semantic Segmentation}} Given an image $I$, and a label set $\mathcal{L} = \{l_0, l_1,\\ \cdots, l_p\}$, where $p$ is the total number of classes and $l_0$ represents the background label. The semantic segmentation task is to assign a label from $\mathcal{L}$ to each pixel in the image $I$. 
In the case of fully supervised setting, the dataset $\mathcal{D}$ consists of images and their corresponding pixel-level class-specific annotations (expensive pixel-level annotations). However, in the {\em weakly-supervised} setting, the dataset consists of images and corresponding annotations that are relatively easy to obtain, such as tags/ labels of objects present in the image. Let us define $Z = \mathcal{L} \backslash l_0$ to be the set of total object labels we are interested in. Thus, the dataset in our case is $\mathcal{D}=\{I_i, {\bf z}_i\}_{i=1}^N$, where ${\bf z}_i \subseteq Z$ are the object labels present in the $i$-th image. The goal thus is to learn the CNN parameters ($\theta$) for the semantic segmentation task using the weak dataset $\mathcal{D}$.
\vspace{-4mm}
\section{Discovering Class-Specific Pixels for Weakly-Supervised Semantic Segmentation}
\label{sec:propApproach}
\vspace{-3mm}

To train a CNN for the semantic segmentation task, we need pixel-level annotations. In the case of weakly-supervised setting, the challenge is to approximate these annotations from image labels and other weak cues such as saliency. To obtain such approximate annotations our approach consists of three main components. First, the fully convolutional attention network (Section~\ref{sec:fcan}) for multiple object categories that allows us to reliably localize objects in an image. Second, mining of salient regions using a simple hierarchical approach (Section~\ref{sec:hierSal}). Third, making use of pixel-level class-specific information obtained using the combination of attention and saliency based cues to guide the training algorithm (Section~\ref{sec:training}). In what follows, we talk about each of these components in detail. 

\begin{figure}[]
\centering
\includegraphics[scale=0.135]{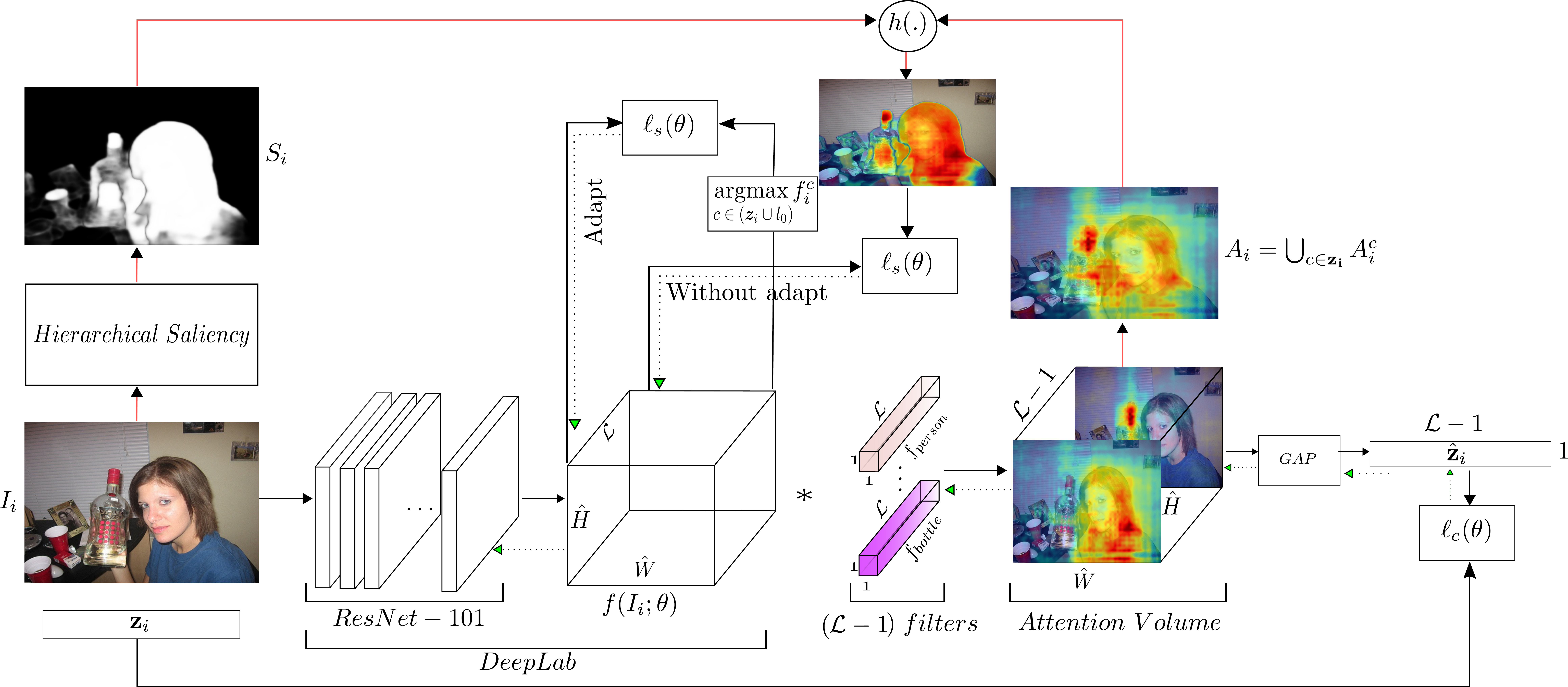}
\caption{A schematic illustration of our proposed approach. We use the same network for both image classification and semantic segmentation tasks. This allows us to obtain attention maps in a fully convolutional manner, without training a new classification network. Arrows with green head represent the backward pass. Refer to Section~\ref{sec:propApproach} for further details.}
\label{fig:propArch}
\end{figure}

\vspace{-4mm}  
\subsection{Fully Convolutional Attention Network}
\label{sec:fcan}
\vspace{-2mm}
It is well known that while classifying an object CNN-based classifiers focus more on certain discriminative areas (pixels) of an object in an image~\cite{zhou2016CAM}. This property of CNNs is extensively utilized by different approaches~\cite{simonyan2013deep, springenberg2014striving, zhou2016CAM} in localizing objects in images. Some of the approaches~\cite{simonyan2013deep, springenberg2014striving} use image gradients to localize objects while others~\cite{zhou2016CAM} use global average pooling (GAP) based classifier architecture. We study the latter approach, and propose a convolutional variant of {\em Class Activation Mapping} (CAM)~\cite{zhou2016CAM}. CAM uses a standard CNN, and just before the final classification layer, it averages the activations across each channel by using a GAP layer. It then passes the averaged activations through a Fully-Connected (FC) layer that produces the final class scores. It can be shown that CAM is essentially taking an inner product between the class-specific weights (FC layer parameters) and the pixel-wise feature vectors (last convolutional feature map) to obtain attention. We propose an FCAN where, instead of FC weights, we use class-specific convolutional filters and push the GAP layer at the end. Specifically, we obtain multi-object attention maps (shown as `Attention Volume' in Figure~\ref{fig:propArch}) by taking the inner product between the class-specific convolutional filters and the penultimate feature volume in the network, followed by averaging the activations using a GAP layer. This allows us to use the segmentation network directly to obtain the attention maps, instead of training a separate classification network. In detail, we re-purpose the fully convolutional segmentation network~\cite{chen2016deeplab} to solve the classification task by adding $|Z|$ additional convolutional filters of size $1\times 1\times K$ to the last layer of the segmentation network, where $K$ is the channel dimension of the last layer of the standard segmentation network (typically $K = |\mathcal{L}|$). Note that, we do not employ convolutional filter for the background as we are interested in only localizing the foreground objects, which in turn can help us find the cues for the background as well. We then add a GAP layer (we find that, Global Max Pooling, as suggested by \citep{oquab2015object}, underestimates the size of the objects) on the last convolution volume to obtain class-specific confidence scores for an image. As in~\cite{oquab2015object}, we treat the multi-label classification problem as $|Z|$ independent binary classification problems to train the network under following objective:
\vspace{-2mm}
\begin{equation} 
\label{eq:classificationLoss}
\ell_c (\theta) = \frac{1}{N} \sum_{i=1}^N - \bar{\bf z}_i\log(\sigma(\hat{{\bf z}}_i)) - (1- \bar{\bf z}_i)\log(1-\sigma(\hat{{\bf z}}_i)) 
\end{equation}
where $\bar{\bf z}_i$, $\hat{{\bf z}}_i$ and $\sigma(.)$ are the ground-truth image-level label vector (`1' if the object is present, otherwise, `0'), network prediction scores and the sigmoid function, respectively. All the operations in equation~(\ref{eq:classificationLoss}) are element-wise. Once the network is trained under this objective, the last convolution volume represents the attention volume $V$ for $|Z|$ categories, as show in the Figure~\ref{fig:propArch}. Then, the attention maps for a given image is obtained as the set of attention maps/slices of the attention volume $V$ corresponding to the object categories ${\bf z}$ present in the image. Formally, we obtain the normalized attentions $A_i$ for the $i$-th image as $A_i=\bigcup_{c\in {\bf z}_i} A_i^c$, where $A_i^c$ represents the normalized attention map for the $i$-th image corresponding to the $c$-th object category. We normalize each slice of the volume $V$ independently between $0$ to $1$ to obtain $A_i^c$. Note that, we use atrous convolutions~\cite{holschneider1990real, chen2016deeplab} to keep the prediction resolution sufficiently large in the last layers of the network, thus, there is no need to calculate the attentions at the earlier layers to get the finer details. 

Although, as explained earlier, the global average pooling forces the CNN to expand the attention maps, this spread is still limited and in some cases even stretches to background pixels as can be seen in the Figure~\ref{fig:hierSal}. In other words, even though the attention maps that we obtain using the fully convolutional approach are quite accurate in locating an object, they are not very precise when it comes to pixel-level localization which is very crucial for obtaining pixel-level class-specific cues for the weakly-supervised segmentation task. In the next section we partially address this issue by combining these attention maps with the class-agnostic saliency maps that we obtain using a simple hierarchical approach.

\vspace{-4.5mm}
\subsection{Hierarchical Saliency For Multiple Salient Objects}
\vspace{-2mm}
\label{sec:hierSal}
One of the major limitations of salient object detectors such as~\cite{liu2016dhsnet}, is that they often fail to detect multiple salient objects in an image. An example of such a case is shown in the Figure~\ref{fig:hierSal}. To address this, we propose a simple hierarchical approach that allows the saliency network to discover new salient regions. In more detail, given a salient region detector~\cite{liu2016dhsnet}, we first find the most salient region by thresholding the output of the saliency detector, and then remove/erase it from the image by replacing its pixel values by the average pixel value over the entire dataset and pass the image with the erased regions again through the saliency detector. Formally, let us denote $S_1$ and $S_2^e$ as the saliency maps of the given image and the image obtained after the first erasing, respectively. Then, we combine $S_1$ and $S_2^e$ to obtain $S_2$ by assigning the maximum saliency score to each pixel $i$ as follows, $S_2(i) = \max (S_1(i), S_2^e(i))$. This allows the saliency detector to discover the next most salient region in the same image. As shown in the Figure~\ref{fig:hierSal} (shown for two erasing steps), this simple approach allows the saliency detector to obtain saliency maps for images containing multiple salient objects. Note that, as opposed to~\cite{wei2017object}, hierarchical saliency detection method does not require the retraining of the network after each erasing and can utilize any off-the-shelf saliency detector without any modifications.

As mentioned earlier, the attention maps give us the class-specific information and corresponding landmark regions for the categories present in the image, whereas, saliency gives us the foreground/background cues. Neither attention nor saliency individually can provide reliable pixel-level class-specific cues. Thus, we combine the attention and saliency maps using a user-defined function $h(.,.)$. Specifically, for a given image, we compute the element-wise harmonic mean (we empirically found it to be better suited than arithmetic or geometric mean) between each category-specific normalized attention map $A^c$ and the saliency map $S$, and obtain the final approximate ground-truth labels using hard thresholding. This procedure is summarized in Algorithm~\ref{alg:combAttnSal}. The user-defined parameter $\gamma$ in Algorithm~\ref{alg:combAttnSal} represents the threshold above which a pixel is assigned to the foreground class. The final localization cues $M$, obtained using this approach, are reliable and remove many false activations as shown in the Figure~\ref{fig:hierSal}. We use these cues to guide the training of the CNN (explained in Section~\ref{sec:training}). 

\begin{algorithm}[t]
\caption{Discovering Class-Specific Pixels}\label{alg:combAttnSal}
\small
\begin{algorithmic}[1]
\INPUT Image Labels ${\bf z}$; Saliency Map $S$; Attention Maps $A$; $\gamma$
    \State $M = zeros(n)$, where $n$ is the number of pixels
	\For{for each $c \in {\bf z}$ and each pixel $m$}
        \State $H(m, c) = h(A^c(m), S(m))$
    \EndFor
	\For{for each pixel $m$}
		\If{$H(m) < \gamma$} \Comment{$H(m)$ has $|{\bf z}|$ elements}
			\State $M(m) = l_0$	\Comment{Assign background}
		\Else
			\State $M(m) = \argmax(H(m))$ \Comment{Assign foreground}
		\EndIf
	\EndFor

\OUTPUT Localization cues or approximate labeling $M$
\end{algorithmic}
\end{algorithm}

\vspace{-4mm}
\subsection{Training}
\vspace{-2mm}
\label{sec:training}
The intuition behind our training objective is driven from a simple fact that in order to solve the weakly-supervised semantic segmentation task the network should also be able to solve the classification task. Therefore, in addition to the segmentation loss $\ell_s$ (pixel-wise cross-entropy), we also add an auxiliary classification loss $\ell_c$ (defined in equation~(\ref{eq:classificationLoss})), to our final objective function. These kinds of auxiliary losses have already been explored in the domain of reinforcement learning~\cite{jaderberg2016reinforcement}. We formally define our training objective as, given an image $I_i$, let us denote $\sigma(f^m(I_i; \theta))$ as the network prediction for the $m$-th pixel consisting of the soft-max probabilities over the labels $\mathcal{L}$ (refer to Figure~\ref{fig:propArch}). Let us denote the approximate ground truth for the $m$-th pixel as $\delta^m_i \in \{0,1\}^{|\mathcal{L}|}$, where $\delta^m_i(l) = 1$ at the $l$-th index belonging to the label category of the $m$-th pixel obtained using the Algorithm~\ref{alg:combAttnSal}. Then, the overall objective function is defined as:
\begin{equation}
\label{eq:finalLoss}
\ell(\theta) = \ell_c(\theta) + \ell_s(\theta)
\end{equation}
where, $\ell_c(\theta)$ is the classification loss (equation~(\ref{eq:classificationLoss})) and $\ell_s(\theta) = \frac{1}{N} \sum_{i=1}^N \sum_{m = 1}^n J(\sigma(f^m(I_i; \theta))\\,\delta_i^m)$. Here, $J(.,.)$ is the pixel-wise cross-entropy loss.
Additionally, we found that as the network trains under both the $\ell_s$ and $\ell_c$, it learns to find even better localization cues as we have additional segmentation loss focusing on pixel-level accuracy. This becomes the basis for our {\em adaptive} training where we iteratively adapt (update) the localization cues after a fixed number of training steps (for example, $10$K). Formally, at the adapt step, the localization cues are obtained as $M_i(m) = \argmax_{c \in {\bf z}_i\cup l_0} f^m_c(I_i; \theta)$. We then continue to train the network under the same objective (equation~(\ref{eq:finalLoss})) with these new cues (refer to the Figure~\ref{fig:propArch}). 

At the test time, we discard the final convolutional layer - meant for the classification task to obtain attention cues - and obtain the segmentation maps from the penultimate layer.

\vspace{-5mm}
\section{Experimental Results, Comparisons and Analysis}
\label{sec:results}
\vspace{-3mm}
We now describe the dataset and the experimental setup (Section~\ref{sec:datasetSetup}), followed by the comparison of our approach with the current state-of-the-art methods to show that our method outperforms all the existing methods on the challenging PASCAL VOC 2012 benchmark (Section~\ref{sec:comparisonSOA}). We then perform some analysis of our approach in Section~(\ref{sec:analysis}) for the purpose of building better understanding of the method.  
\vspace{-3mm}
\subsection{Dataset and Experimental Setup}
\label{sec:datasetSetup}
\vspace{-2mm}
\paragraph{\textcolor{bmv@sectioncolor}{Dataset}} We evaluate our framework on the challenging PASCAL VOC12 segmentation benchmark dataset~\cite{everingham2010pascal}, that contains 20 foreground object categories and one background category. The original dataset contains $1,464$ training images. Following common practice \cite{chen2014semantic, hariharan2011semantic, pinheiro2015image}, we augment the dataset with the extra annotations provided by \cite{hariharan2011semantic}. This gives us total of $10,582$ training images. The validation and test sets contain $1,449$ and $1,456$ images, respectively. No additional data is being used in the entire train/test pipeline.

\vspace{-4mm}
\paragraph{\textcolor{bmv@sectioncolor}{Saliency Network}} We employ DHSNet \cite{liu2016dhsnet} as the saliency detector, and use our hierarchical approach (Section~\ref{sec:hierSal}) to allow it to discover different salient regions which is useful in situations when images contain multiple salient objects. For the first erasing, any pixel with the saliency score greater than $0.7$ is erased from the image and replaced with the average pixel value. Similarly, a threshold of $0.8$ is used for the second erasing.

\vspace{-4mm}
\paragraph{\textcolor{bmv@sectioncolor}{Unified Attention and Segmentation Network}} Our unified network is based on DeepLab-V2 \cite{chen2016deeplab} whose parameters are initialized by the ResNet-101 \cite{he2016deep} pretrained on ImageNet \cite{deng2009imagenet} for the classification task. We use Tensorflow \cite{abadi2016tensorflow} to implement the Deeplab \footnote{The code is available at \url{https://github.com/arslan-chaudhry/dcsp_segmentation}}. We append $20$, $1\times1\times21$ convolution filters at the last layer of the segmentation network. The weights of the last two layers are initialized with the Gaussian having zero mean, $0.01$ standard deviation and biases with zeros. The CNN hyper-parameters used are: momentum ($0.9$), weight decay ($0.0005$), batch size ($10$). The initial learning rate is set to $0.001$ and then `poly' (with $10K$ maximum iterations) learning rate policy is deployed as suggested by \cite{chen2016deeplab}. We randomly crop the images to $321 \times 321$ and also perform random scaling and mirroring. In order to obtain our first attention network, we use PASCAL VOC 2012 images to train the above defined network for $30K$ iterations under the {\em classification objective} as defined in equation~(\ref{eq:classificationLoss}). We, then, train the network for $10K$ iterations optimizing the objective defined in equation~(\ref{eq:finalLoss}) followed by updating/adapting the ground-truth cues and retraining the network for another $10K$ iterations minimizing the same objective. Note that, the learning rate is reset to $0.001$ after the \emph{adapt} step. Adapting further does not improve results as the network has already saturated the pixel-level cues obtained from weak image labels. The background threshold $\gamma$ (see Algorithm~\ref{alg:combAttnSal}) is set to $0.4$. Given that the saliency maps are already obtained from the off-the-shelf saliency detector~\cite{liu2016dhsnet}, the complete training framework is end-to-end trainable.

At test time, we calculate the feature maps at three different scales $(1,\ 0.75,\ 0.5)$ and fuse them by taking maximum at each location to obtain the final prediction. 

\begin{table}[!t]
\resizebox{0.35\textwidth}{!}{\begin{minipage}[t]{.4\textwidth}
\begin{threeparttable}
    \caption{Comparison: Weakly-Supervised Semantic Segmentation Methods on PASCAL VOC12. $^{1}$Uses $40$K additional images from Flickr. $^2$Depends on MCG~\cite{arbelaez2014multiscale} which requires pixel-level supervision. $^3$. Uses ResNet-101 in saliency network. $^4$. Reuses ImageNet images for segmentation task. Also, manuscript is unpublished/ not peer-reviewed. $^5$ Based on ResNet-101 whereas few other methods use VGG-16~\cite{Simonyan15}.}
\label{tab:compareSOA}
\begin{tabular}{@{}cccc@{}}
\toprule
Methods                                                 & CRF & \begin{tabular}[c]{@{}c@{}}mIOU\\ (Val)\end{tabular} & \begin{tabular}[c]{@{}c@{}}mIOU\\ (Test)\end{tabular} \\ \midrule
EM-Adapt~\cite{papandreou2015weakly} 
									& \cmark & 38.2\% & 39.6\% \\ 
									\midrule
                                    	& \xmark & 33.3\% & 35.6\% \\ 
                                    	\cmidrule(l){2-4} 
\multirow{-2}{*}{CCNN~\cite{pathak2015constrained}}
                                  	& \cmark & 35.3\% & - \\ 
                                  	\midrule
                                     & \xmark & 44.3\% & - \\ 
                                     \cmidrule(l){2-4} 
\multirow{-2}{*}{SEC~\cite{kolesnikov2016seed}}                                   
									& \cmark & 50.7\% & 51.7\% \\ 
									\midrule
STC~\cite{wei2016stc} \tnote{1}                                                     
									& \cmark & 49.8\% & 51.2\%  \\ 
									\midrule
MIL~\cite{pinheiro2015image} \tnote{2}                                                 												& \xmark & 42.0\% & 40.6\%  \\ 
							\midrule
                                     & \xmark & 50.4\% & 50.6\% \\ 
                                     \cmidrule(l){2-4} 
\multirow{-2}{*}{AugFeed~\cite{qi2016augmented} \tnote{2}}                           									& \cmark & 54.3\%  & 55.5\% \\ 
									\midrule
Combining Cues~\cite{RoyWeakly2017}
									& \cmark & 52.8\%  & 53.7\%  \\
									\midrule
AE-PSL~\cite{wei2017object}
									& \cmark & 55.0\%  & 55.7\%  \\ 
									\midrule
									& \xmark & 51.2\% & - \\
									\cmidrule(l){2-4} 
\multirow{-2}{*}{Joon et al.~\cite{joon17cvpr} \tnote{3}}
									& \cmark & 55.7\% & 56.7\% \\
									\midrule
									& \xmark & 56.9\% & 57.7\% \\
									\cmidrule(l){2-4} 
\multirow{-2}{*}{Mining Pixels~\cite{HouMining2016} \tnote{4}}
									& \cmark & 58.7\% & 59.6\% \\
									\midrule
									& \xmark & 56.5\% & 57.04\% \\ 
                                     \cmidrule(l){2-4}
\multirow{-2}{*}{\textbf{DCSP-VGG16} (ours)} 
									& \cmark & 58.6\% & 59.24\% \\
									\cmidrule(l){2-4}
									& \xmark  & 59.5\% & 60.3\% \\ 
									\cmidrule(l){2-4} 
\multirow{-2}{*}{\textbf{DCSP-ResNet-101} (ours) \tnote{5}} 
									& \cmark & \textbf{60.8\%} & \textbf{61.9\%} \\
			 						\bottomrule
\end{tabular}
\end{threeparttable}
\end{minipage}}
\hspace{2.5cm}
\resizebox{0.29\textwidth}{!}{\begin{minipage}[t]{.4\textwidth}
\centering
\begin{threeparttable}
\caption{Ablation analysis of our approach on PASCAL VOC12 {\em val}. We train the network for $10K$ iterations, then adapt the attention cues followed by training for another $10K$ iterations. (All the results are with ResNet-101 unless stated otherwise.)}
\label{tab:ablAnalysis}
\begin{tabular}{cccl}
\toprule
\multicolumn{1}{l}{CRF} & \begin{tabular}[c]{@{}c@{}}Saliency\\ Mask\end{tabular} & Adapt & mIOU                       \\ \hline
\multirow{4}{*}{\xmark}      & \multirow{2}{*}{$S_1$}                                     & \xmark                                                    & \multicolumn{1}{c}{55.4\%} \\ \cline{3-4} 
                        &                                                         & \cmark                                                    & 55.7\%                     \\ \cline{2-4} 
                        & \multirow{2}{*}{$S_3\ (HS)$}                                     & \xmark                                                    & 58.5\%                     \\ \cline{3-4} 
                        &                                                         & \cmark                                                    & 59.5\%                     \\ \hline
\multirow{4}{*}{\cmark}      & \multirow{2}{*}{$S_1$}                                     & \xmark                                                    & 56.0\%                     \\ \cline{3-4} 
                        &                                                         & \cmark                                                    & 56.3\%                     \\ \cline{2-4} 
                        & \multirow{2}{*}{$S_3\ (HS)$}                                     & \xmark                                                    & 60.4\%                     \\ \cline{3-4} 
                        &                                                         & \cmark                                                    & 60.8\%                     \\ \hline
\end{tabular}
\end{threeparttable}
\begin{threeparttable}
\caption{Effects of {\em Hierarchical Saliency}: Notice how missing objects are being segmented when trained using $S_3$ (examples are from `val' dataset).}
\setlength\tabcolsep{1.5pt}
\begin{tabular}{|c|c|c|}
\toprule
\multicolumn{1}{c}{\includegraphics[scale=0.1]{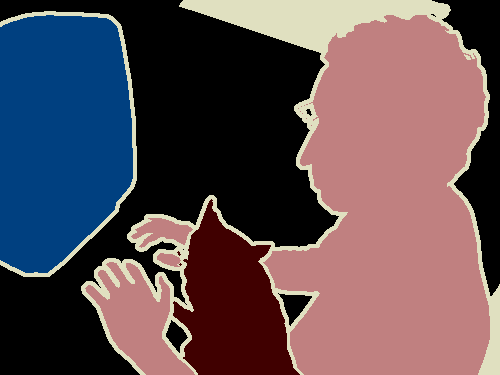}} & 
\multicolumn{1}{c}{\includegraphics[scale=0.1]{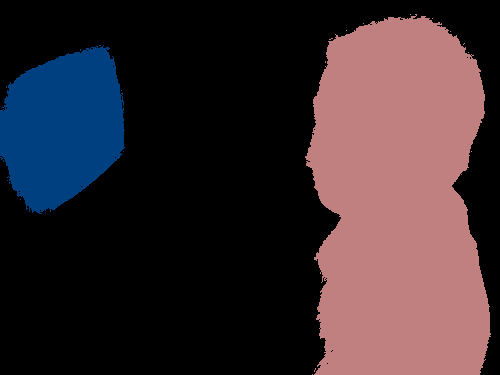}} & 
\multicolumn{1}{c}{\includegraphics[scale=0.1]{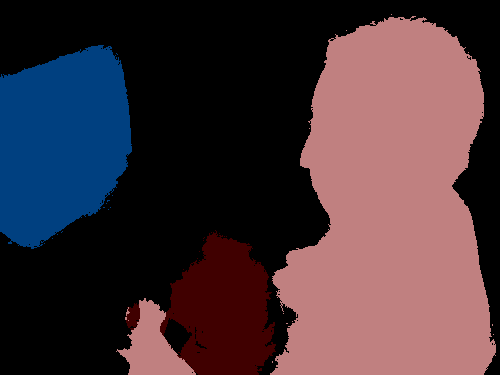}} \\

\multicolumn{1}{c}{\includegraphics[scale=0.1]{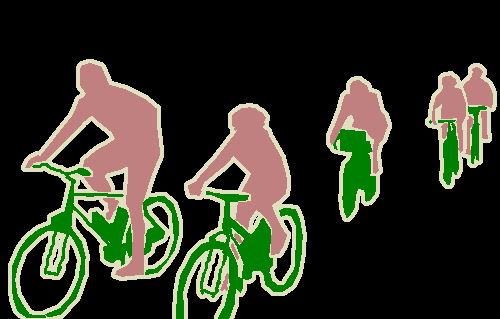}} & 
\multicolumn{1}{c}{\includegraphics[scale=0.1]{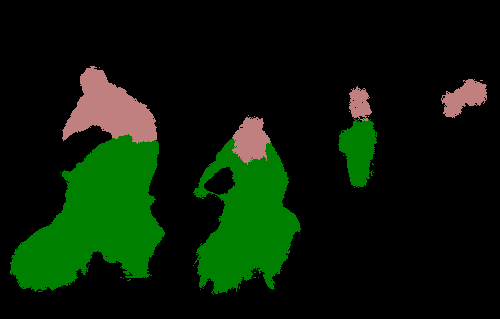}} & 
\multicolumn{1}{c}{\includegraphics[scale=0.1]{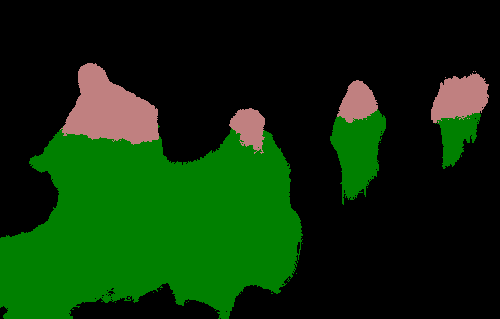}} \\

\multicolumn{1}{c}{\includegraphics[scale=0.1]{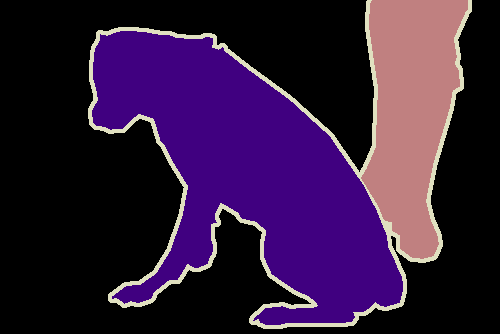}} & 
\multicolumn{1}{c}{\includegraphics[scale=0.1]{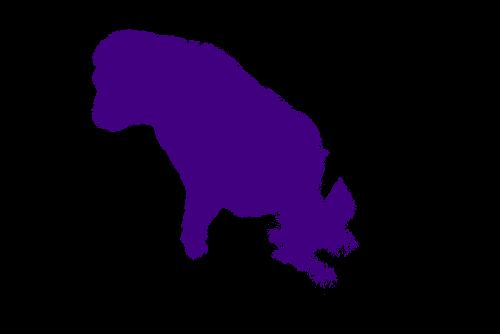}} & 
\multicolumn{1}{c}{\includegraphics[scale=0.1]{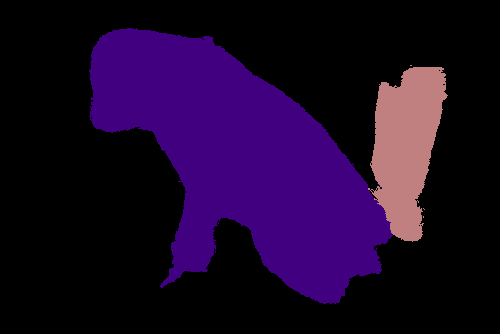}} \\

\multicolumn{1}{c}{\includegraphics[height=1.8cm, width=1.8cm]{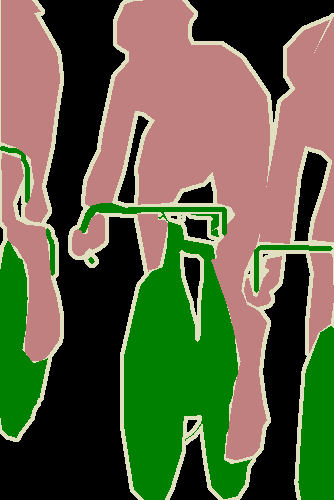}} & 
\multicolumn{1}{c}{\includegraphics[height=1.8cm, width=1.8cm]{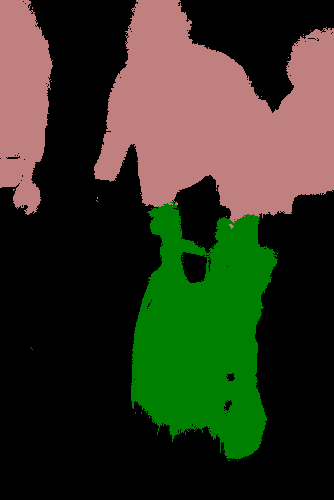}} & 
\multicolumn{1}{c}{\includegraphics[height=1.8cm, width=1.8cm]{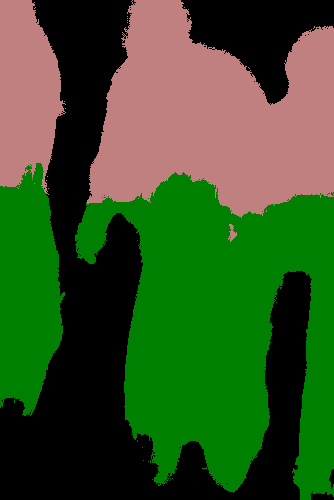}} \\

\multicolumn{1}{c}{Ground Truth}& \multicolumn{1}{c}{$S_1$} & \multicolumn{1}{c}{$S_3\ (HS)$}
\end{tabular}
\label{tab:abalSegmentations}
\end{threeparttable}

\hspace{1mm}
\caption{Effects of jointly training the saliency and attention cues in a unified segmentation network (results are from 'val' dataset).}
\begin{tabular}{cc}
\toprule
Joint Training & mIOU   \\ \hline
\xmark              & 48.3\% \\ \hline
\cmark              & 59.5\% \\ \hline
\end{tabular}
\label{tab:abalJointTraining}
\end{minipage}}

\end{table}
\vspace{-5mm}
\subsection{Comparison with State-of-the-arts}
\label{sec:comparisonSOA}
\begin{figure}[t]
\centering
\setlength\tabcolsep{5pt}
\scalebox{0.81}{
\begin{tabular}{llllll}
 \multicolumn{1}{c}{\includegraphics[scale=0.16]{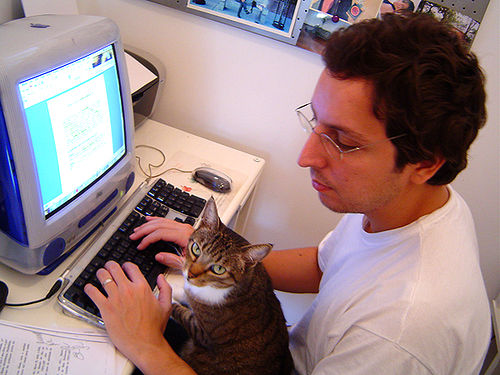}}&  
 \multicolumn{1}{c}{\includegraphics[scale=0.12]{images/gt/2010_004479.png}}&  
 \multicolumn{1}{c}{\includegraphics[scale=0.12]{images/supplementary_fig_adaptive/crf/adpat2/s3/2010_004479.png}}&  
 
 \multicolumn{1}{c}{\includegraphics[scale=0.12]{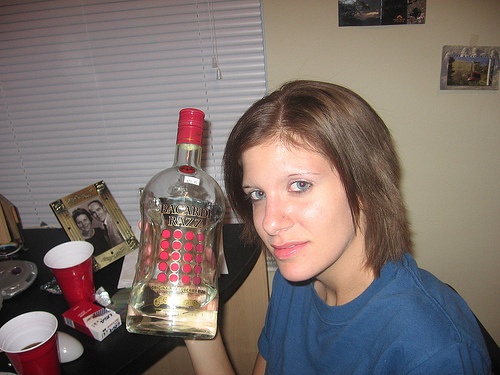}}&  
 \multicolumn{1}{c}{\includegraphics[scale=0.12]{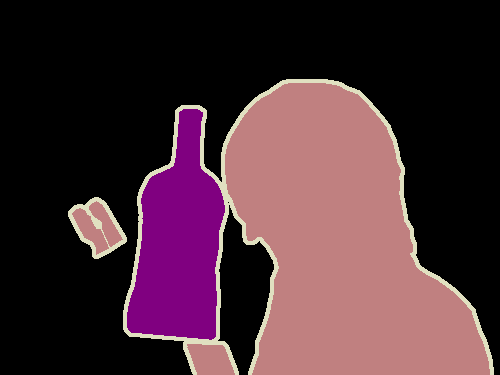}}&  
 \multicolumn{1}{c}{\includegraphics[scale=0.12]{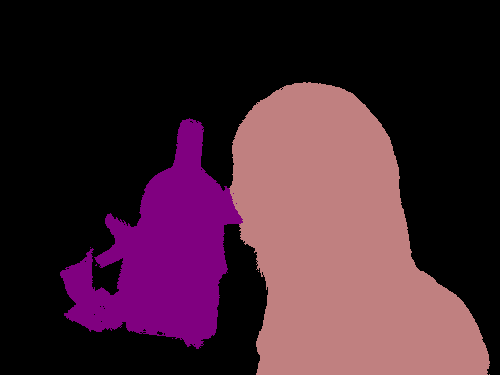}} \\ 

 \multicolumn{1}{c}{\includegraphics[scale=0.12]{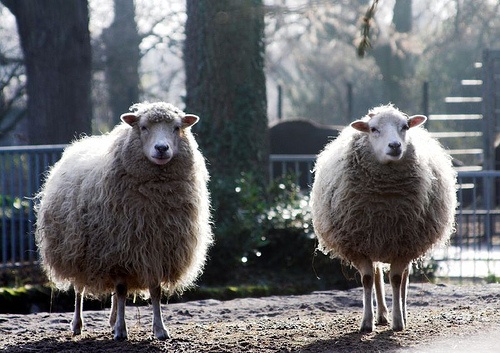}}&  
 \multicolumn{1}{c}{\includegraphics[scale=0.12]{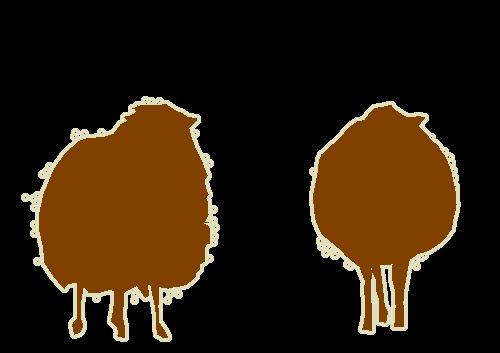}}&  
 \multicolumn{1}{c}{\includegraphics[scale=0.12]{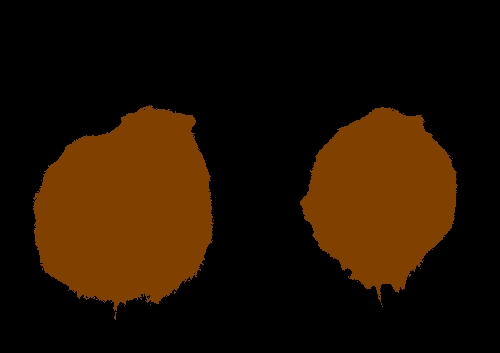}}&  
 
 \multicolumn{1}{c}{\includegraphics[scale=0.12]{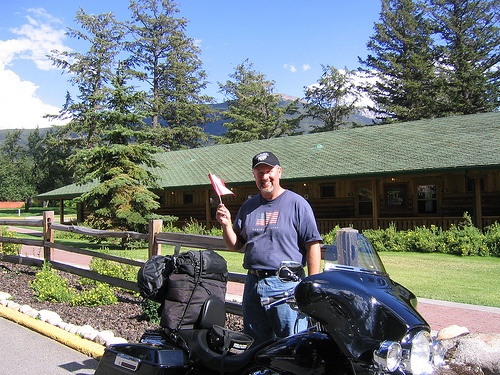}}&  
 \multicolumn{1}{c}{\includegraphics[scale=0.12]{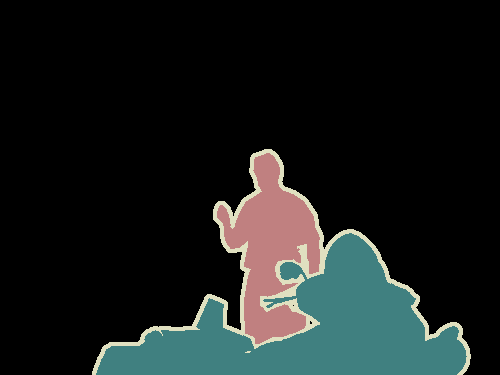}}&  
 \multicolumn{1}{c}{\includegraphics[scale=0.12]{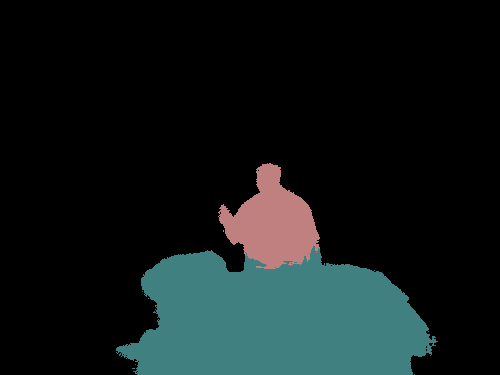}}  \\  
 
 \multicolumn{1}{c}{\includegraphics[scale=0.12]{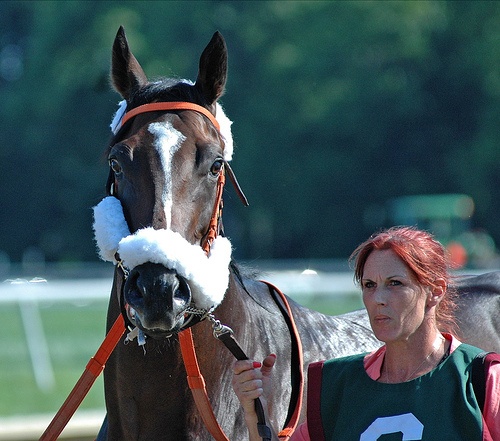}}&  
 \multicolumn{1}{c}{\includegraphics[scale=0.12]{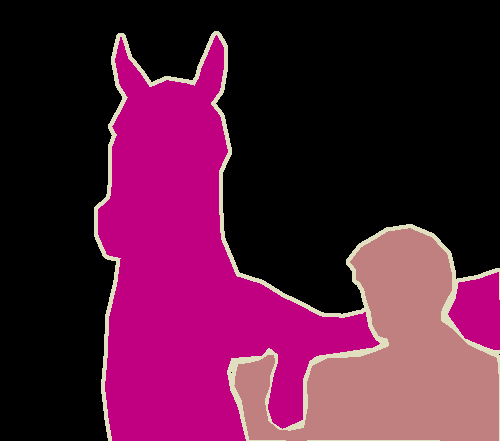}}&  
 \multicolumn{1}{c}{\includegraphics[scale=0.12]{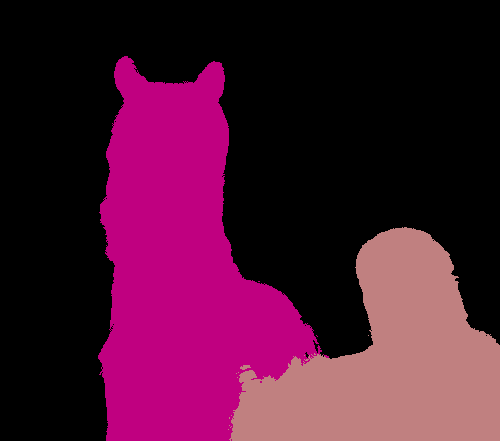}}&  
 
 \multicolumn{1}{c}{\includegraphics[scale=0.12]{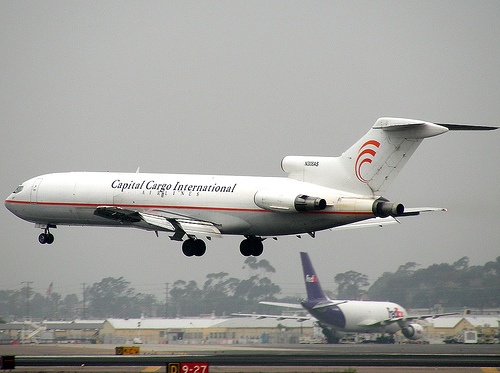}}&  
 \multicolumn{1}{c}{\includegraphics[scale=0.12]{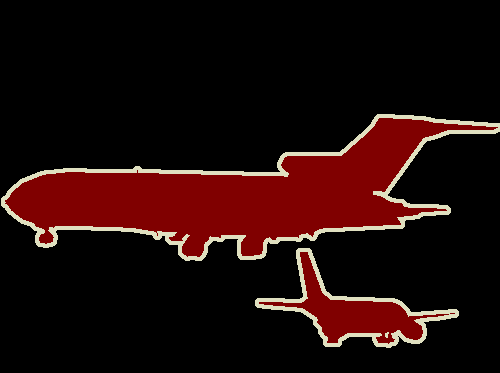}}&  
 \multicolumn{1}{c}{\includegraphics[scale=0.12]{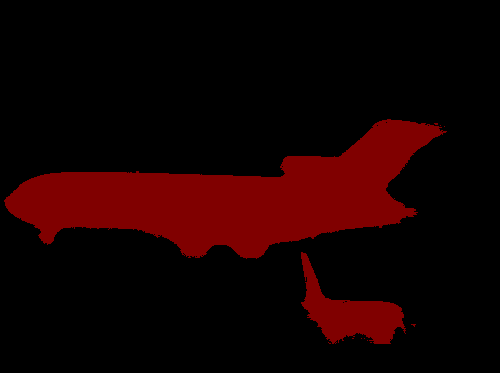}}  \\
 
 \multicolumn{1}{c}{\includegraphics[height=2cm, width=2cm]{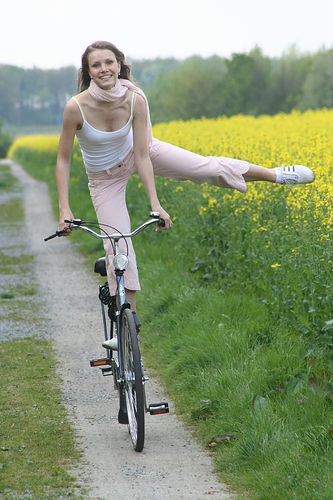}}&  
 \multicolumn{1}{c}{\includegraphics[height=2cm, width=2cm]{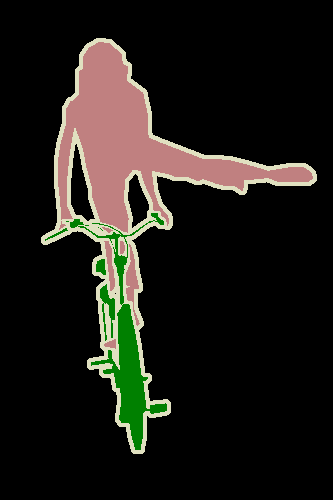}}&  
 \multicolumn{1}{c}{\includegraphics[height=2cm, width=2cm]{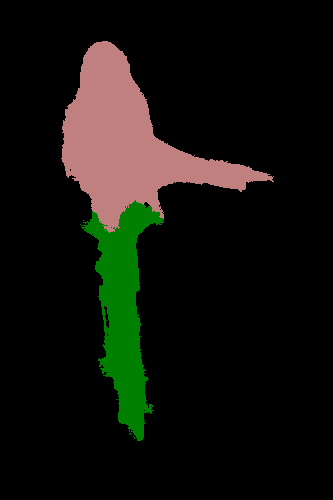}}&
  
 \multicolumn{1}{c}{\includegraphics[height=2cm, width=2cm]{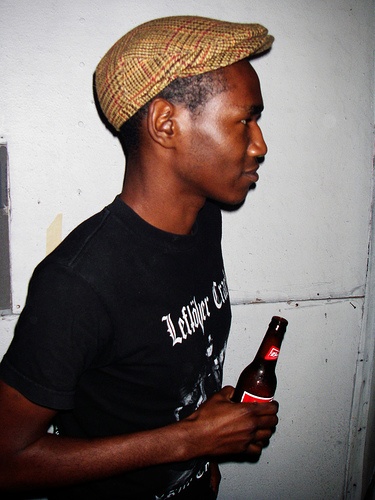}}&  
 \multicolumn{1}{c}{\includegraphics[height=2cm, width=2cm]{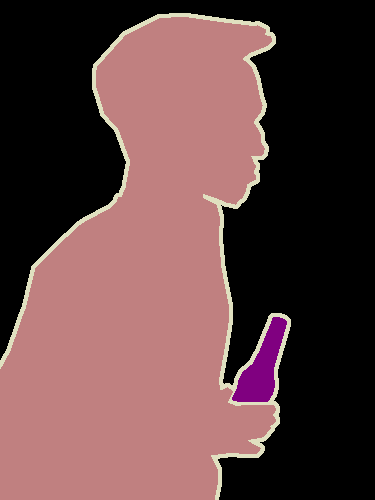}}&  
 \multicolumn{1}{c}{\includegraphics[height=2cm, width=2cm]{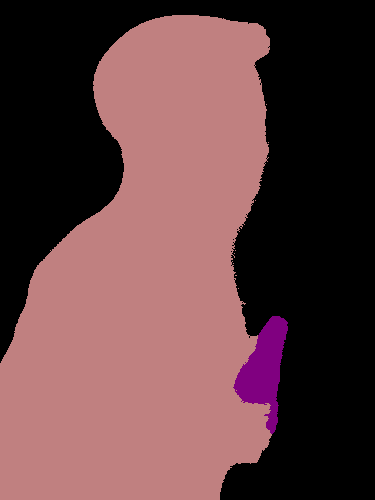}} \\  
  
 \multicolumn{1}{c}{\includegraphics[scale=0.12]{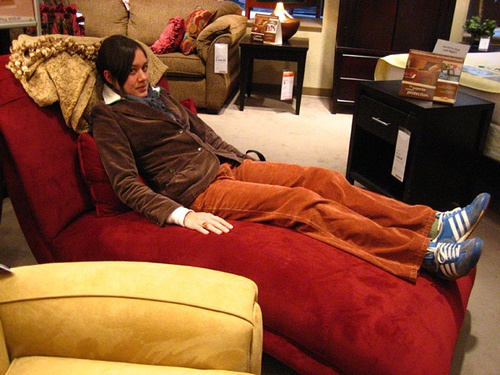}}&  
 \multicolumn{1}{c}{\includegraphics[scale=0.12]{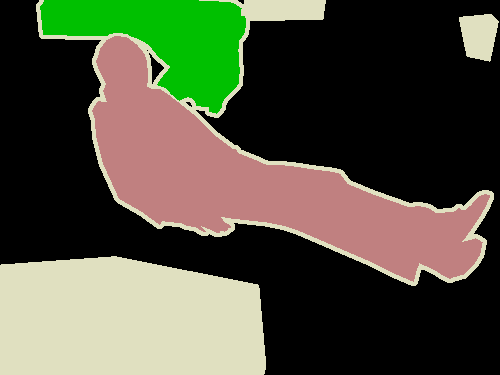}}&  
 \multicolumn{1}{c}{\includegraphics[scale=0.12]{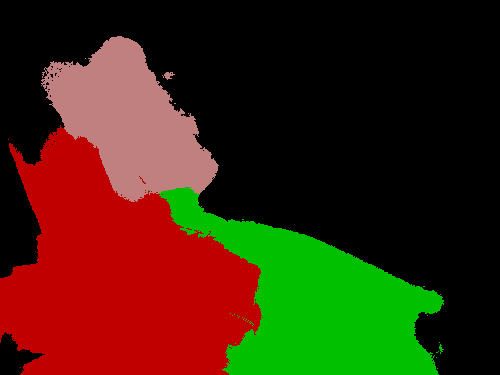}}&  
 
 \multicolumn{1}{c}{\includegraphics[scale=0.12]{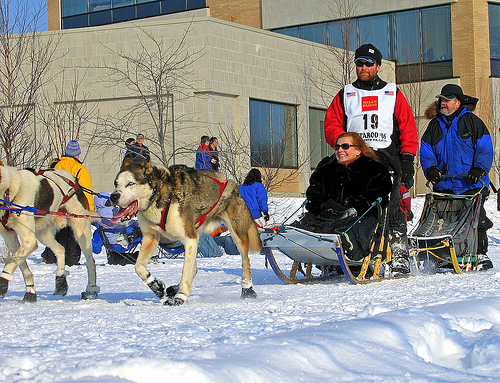}}&  
 \multicolumn{1}{c}{\includegraphics[scale=0.12]{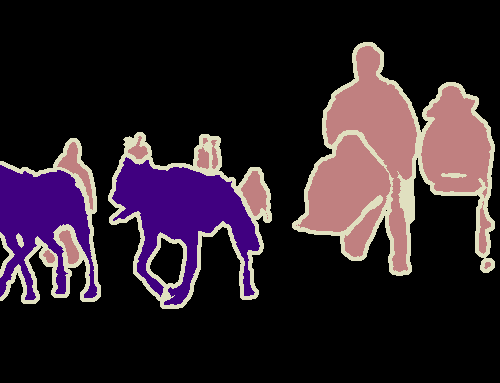}}&  
 \multicolumn{1}{c}{\includegraphics[scale=0.12]{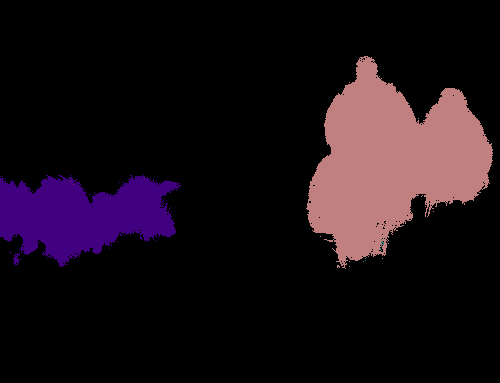}} \\  
 
 \multicolumn{1}{c}{Image}&  \multicolumn{1}{c}{Ground Truth}&  \multicolumn{1}{c}{Our}&  \multicolumn{1}{c}{Image}&  \multicolumn{1}{c}{Ground Truth}&  \multicolumn{1}{c}{Our}
\end{tabular}}
    \caption{Few qualitative results on the PASCAL VOC 2012 val set. As can be seen that the network is able to discover multiple objects and also keeps the boundaries of the objects intact. Bottom row: Two failure cases where the network fails under sever occlusion.}
\label{fig:segmentationExamples}
\end{figure}

\vspace{-2mm}
We compare our method (DCSP) with the existing state-of-the-art weakly-/semi-supervised semantic segmentation approaches. Table~\ref{tab:compareSOA} shows all the comparisons and Figure~\ref{fig:segmentationExamples} shows segmentation visualizations using our approach. From the results in Table~\ref{tab:compareSOA}, we can verify that our simple approach outperforms the existing approaches to weakly-supervised semantic segmentation task on both `val' and `test' sets, thereby, setting the new state-of-the-art. Particularly, the performance gains on the {\em published} state-of-the-art method of Joon et al.~\cite{joon17cvpr} are $\bf{5.1\%}$ and $\bf{5.2\%}$ on `val' and `test' sets, respectively. To highlight the fact that the gains of our proposed approach are not trivial due to the architectural differences (VGG16 vs ResNet-101), we also report results in Table~\ref{tab:compareSOA} with VGG16 variant of our model and still maintain the state-of-the-art performance compared to the published results. 


A few of the methods we compare with depends on stronger supervisions such as scribbles, bounding boxes, MCG~\cite{arbelaez2014multiscale} and spots~\cite{papandreou2015weakly, qi2016augmented, bearman2016s}. In terms of dependencies, along with image labels, our method uses the saliency network (similar to~\cite{HouMining2016,wei2017object}) that is trained on class-agnostic salient region masks, so, once trained, the saliency network does not require retraining for new object categories. Whereas, among the baselines, STC~\cite{wei2016stc} uses additional data ($50K$ Flickr images) for training. Likewise, Mining Pixels~\citep{HouMining2016} reuses the $24K$ ImageNet images along with PASCAL for the segmentation task. Similarly, AugFeed~\cite{qi2016augmented} employs MCG~\cite{arbelaez2014multiscale} generator which is trained using a fully-supervised dataset (pixel-level annotation) and, hence, makes use of stronger supervision. Even without these stronger supervisions, our method consistently outperforms all these baselines. 

The most directly comparable methods to our approach, in terms of supervision and additional dependencies, are AE-PSL~\cite{wei2017object} and~\cite{joon17cvpr}. Both of these methods, like ours, make use of image-level tags of only PASCAL VOC dataset for training. AE-PSL~\cite{wei2017object} requires retraining of classification network after each erasing using the attention cues, whereas our method does not need to retrain the saliency detector.
This renders our training regime simple and efficient. Likewise the performance of~\cite{joon17cvpr} deteriorates significantly (from 55.7\% to 51.2\%) in the absence of CRF post-processing whereas we maintain the competitive performance even without CRF post-processing (60.8\% to 59.5\%).   



\vspace{-4mm}
\subsection{Analysis}
\label{sec:analysis}
\vspace{-2mm}
In Table~\ref{tab:ablAnalysis}, we report how the category-specific pixel discovery obtained by combining the hierarchical saliency along with attention maps improve the results. As shown in the table, \emph{Hierarchical Saliency} ($S_3$, fig~\ref{fig:hierSal}) results in ${4.5 \%}$ gain in mIOU compared to when it is not being used ($S_1$, fig~\ref{fig:hierSal}). We also validate this fact qualitatively in Table~\ref{tab:abalSegmentations} where it can be seen that the hierarchical saliency allows us to semantically segment multiple objects which otherwise would be ignored. Additionally, we also show in Table~\ref{tab:ablAnalysis} that adapting the localization cues as training progresses removes many false positives, thereby, increasing the prediction accuracy. The qualitative gains of adaptive training and different erasing steps in the hierarchical saliency approach are further discussed in the {\em supplementary material}.

In Table~\ref{tab:abalJointTraining} we discuss the benefit of jointly training the saliency and attention cues in a unified segmentation network. Experimentally, we observe performance gains of ${11.2 \%}$ in mIOU as a result of joint training. Intuitively, without joint training, the final segmentation would be an arithmetic combination of attention and saliency maps, trained separately. Once trained jointly, we learn a set of parameters that are specific to the combined task and a shared feature space that generalized well for the segmentation objective than using different feature space mappings for attention and saliency. 
 Additionally, saliency detector trains parameters on class-agnostic masks, whereas, segmentation is a class-specific task, hence, joint training respects the nature of the segmentation objective.

%
%
\vspace{-5mm}
\section{Conclusion and Future Work}
\label{sec:conclusion}
\vspace{-3mm}
We proposed a class-specific pixel discovery method for weakly-supervised semantic segmentation. We showed that properly combining class-specific attention cues (FCAN) with the class-agnostic saliency maps (Hierarchical Saliency) enabled us to reliably obtain pixel-level class-specific cues to improve the performance of the weakly supervised segmentation task. We showed the efficacy of our approach using extensive experiments and reported new state-of-the-art results on PASCAL VOC 2012 dataset.    

One major limitation of the weakly-supervised methods is their inability to detect object boundaries under sever occlusion. This limitation is due to the weak nature of cues that are used to train such methods. To mitigate this shortcoming, an interesting future direction would be to explore the edge and shape-based priors in these methods.
\vspace{-3mm}
\subsection*{Acknowledgements} This work was supported by The Rhodes Trust, EPSRC, ERC grant ERC-2012-AdG 321162-HELIOS, EPSRC grant Seebibyte EP/M013774/1 and EPSRC/MURI grant EP/N019474/1.

\section*{Supplementary Material}
We further analyse the efficacy of our approach \emph{DCSP} that combines the fully convolutional attention maps with the hierarchical saliency masks to obtain reliable pixel-level class-specific cues for the weakly-supervised semantic segmentation task (see Figure 2 in the main paper). Particularly, we compare the performance of the network at different erasing steps and the effect of adapting the localization cues during training.
\section*{Performance Analysis at Different Erasing Steps}
In Tables~\ref{tab:catgValMIOU} and~\ref{tab:catgTestMIOU} we compare the performance gains that we achieve on the weakly-supervised semantic segmentation task using different erasing steps on PASCAL VOC 2012 \emph{val} and \emph{test} sets, respectively. It can be seen from the tables that we get a significant performance boost after the first erasing ($S_2$). The network performance, however, remains consistent (albeit with a small gain) after the second erasing ($S_3$). This could be because, although the PASCAL dataset contains complex images of multiple object categories, the number of salient objects, on average, still remains small. Hence, in most cases $S_2$ would be sufficient to discover the multiple salient objects in the images saturating the network performance on the task.

Another observation from the tables is the poor segmentation accuracy on the categories like \emph{chair}, \emph{table}, \emph{sofa} etc. For example, in Table~\ref{tab:catgTestMIOU} the IOUs for \emph{chair} are $14.9\%$, $20.9\%$ and $20.8\%$, whereas for \emph{aeroplane} these are $73.7\%$, $75.6\%$ and $79.4\%$, after $S_1$, $S_2$ and $S_3$, respectively. Even though the erasing steps are helping us to improve over these categories, the final accuracy is still not satisfactory. Note that, even in the case of a full pixel-level supervision, the IOU on these categories is worse than the other categories ($30.7\%$ for \emph{chair} compared to $84.4\%$ for \emph{aeroplane})~\cite{papandreou15weak}. We suspect that this is due to the elongated nature of the shapes of these categories. For example, in the case of \emph{chair} a large amount of pixels are assigned to its elongated legs and failing to localize these regions will incur a significant performance penalty. Additionally, these categories often appear under sever occlusion and thus, do not maintain a contiguous shape. Since our method approximates the localization cues by combining the attention and saliency, we are always susceptible to the ground-truth cues not having the contiguous regions. Hence, objects that often appear as the set of disjointed regions will not be properly segmented out by our method. Note that, this issue is rampant in most of the existing weakly-supervised semantic segmentation methods~\citep{HouMining2016, wei2017object, kolesnikov2016seed}. To rectify this, one possible solution would be to use edge- and shape-based priors that could localize these elongated and disjointed regions resulting in a better segmentation accuracy for such categories.

\section*{Adaptive Training}
\begin{figure}[htb]
\centering
\caption{Qualitative comparison of using \emph{adaptive} training. Images are taken from PASCAL VOC12 \emph{val} set and are post-processed with the CRF.}
\setlength\tabcolsep{1.5pt}
\begin{tabular}{|c|ccc|ccc|}
\multicolumn{1}{c}{\includegraphics[scale=0.1]{images/gt/2007_000346.png}} & 
\multicolumn{1}{c}{\includegraphics[scale=0.1]{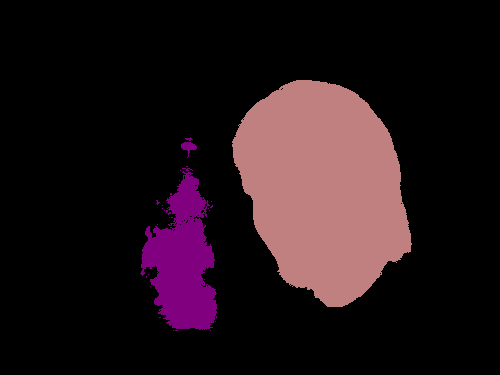}} & 
\multicolumn{1}{c}{\includegraphics[scale=0.1]{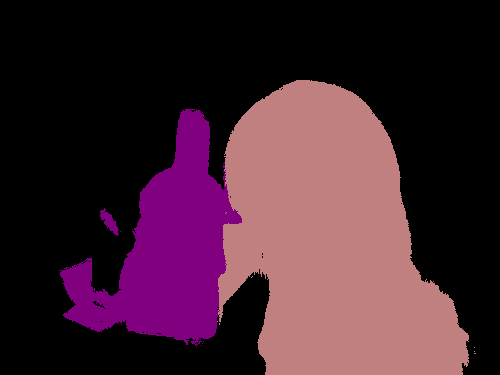}} &
\multicolumn{1}{c}{\includegraphics[scale=0.1]{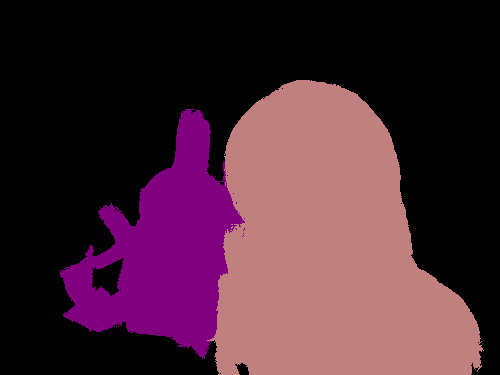}} & 

\multicolumn{1}{c}{\includegraphics[scale=0.1]{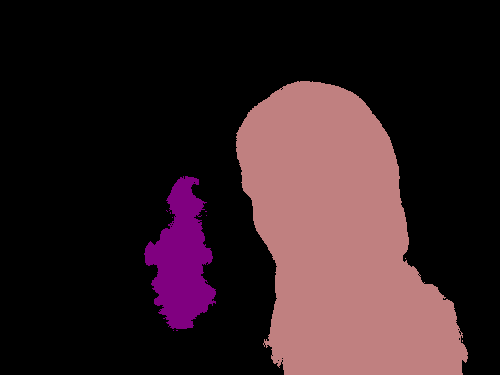}} & 
\multicolumn{1}{c}{\includegraphics[scale=0.1]{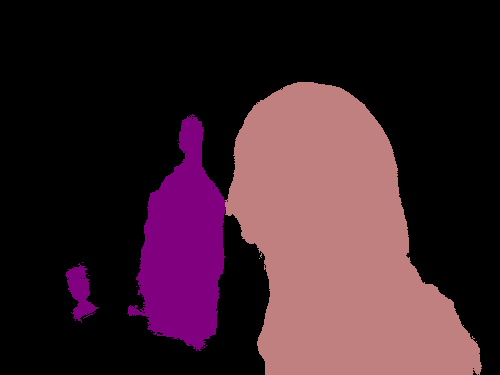}} &
\multicolumn{1}{c}{\includegraphics[scale=0.1]{images/supplementary_fig_adaptive/crf/adpat2/s3/2007_000346.png}} \\

\multicolumn{1}{c}{\includegraphics[scale=0.1]{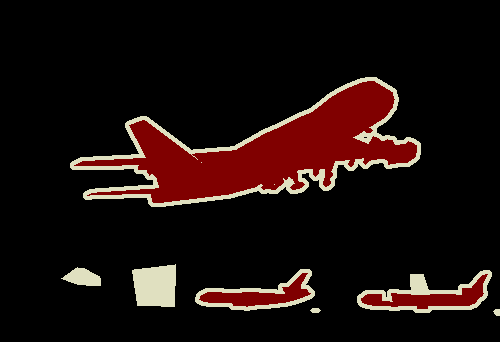}} &
\multicolumn{1}{c}{\includegraphics[scale=0.1]{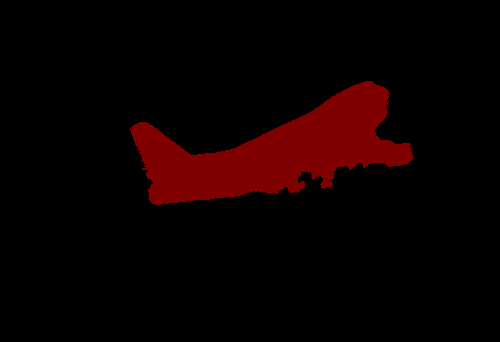}} & 
\multicolumn{1}{c}{\includegraphics[scale=0.1]{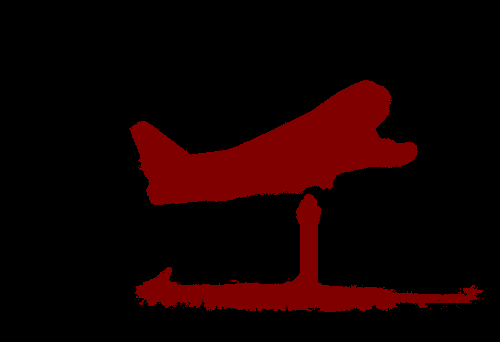}} &
\multicolumn{1}{c}{\includegraphics[scale=0.1]{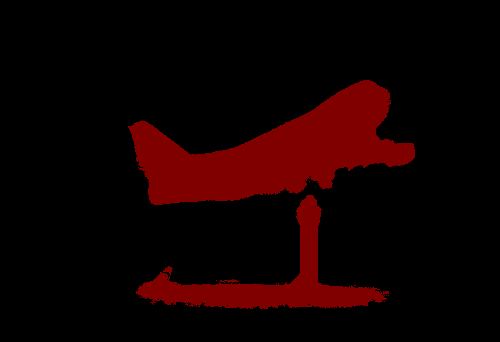}} & 

\multicolumn{1}{c}{\includegraphics[scale=0.1]{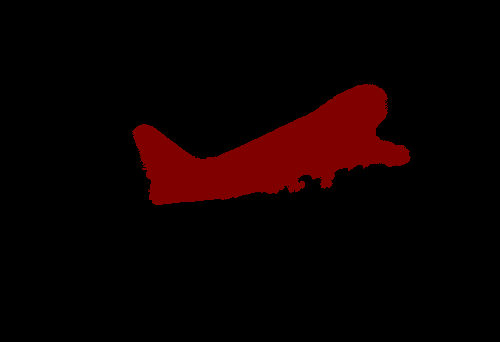}} & 
\multicolumn{1}{c}{\includegraphics[scale=0.1]{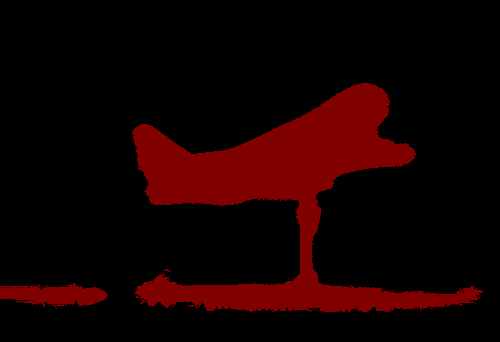}} &
\multicolumn{1}{c}{\includegraphics[scale=0.1]{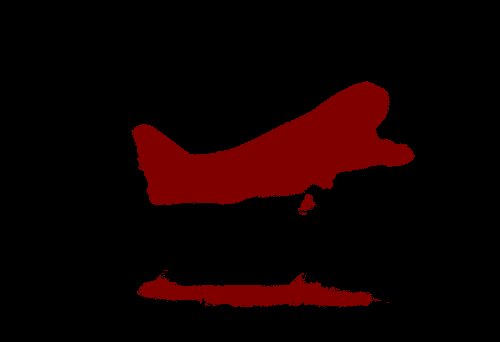}} \\

\multicolumn{1}{c}{\includegraphics[scale=0.1]{images/gt/2007_002823.png}} &
\multicolumn{1}{c}{\includegraphics[scale=0.1]{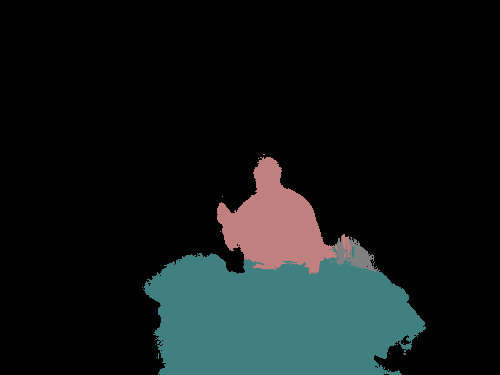}} & 
\multicolumn{1}{c}{\includegraphics[scale=0.1]{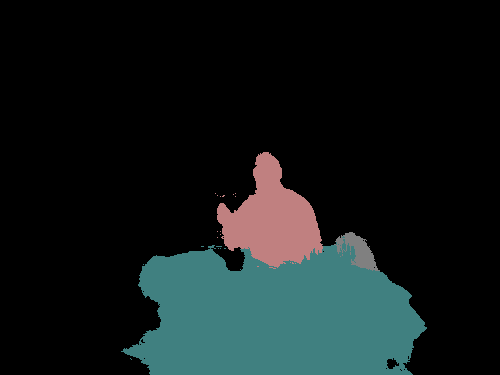}} &
\multicolumn{1}{c}{\includegraphics[scale=0.1]{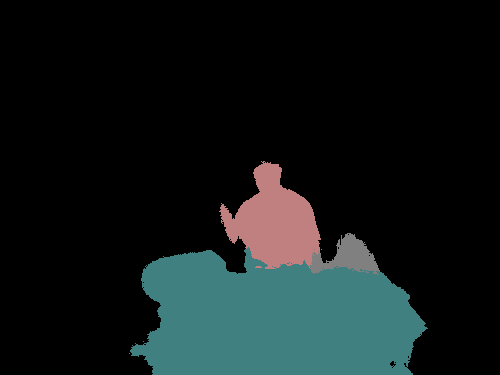}} & 

\multicolumn{1}{c}{\includegraphics[scale=0.1]{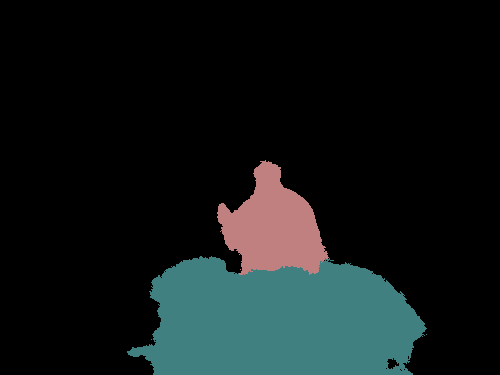}} & 
\multicolumn{1}{c}{\includegraphics[scale=0.1]{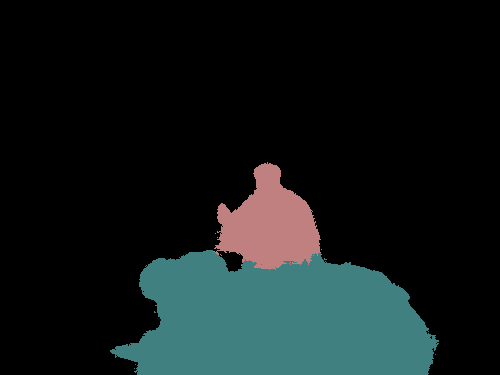}} &
\multicolumn{1}{c}{\includegraphics[scale=0.1]{images/supplementary_fig_adaptive/crf/adpat2/s3/2007_002823.png}} \\

\multicolumn{1}{c}{\includegraphics[scale=0.1]{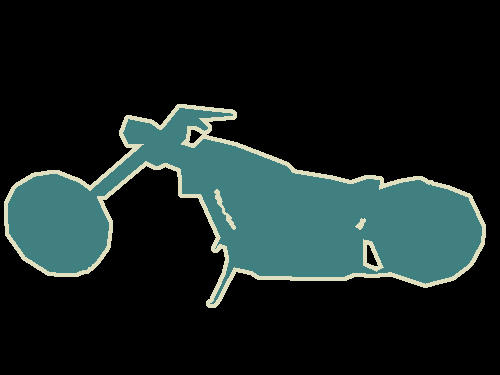}} &
\multicolumn{1}{c}{\includegraphics[scale=0.1]{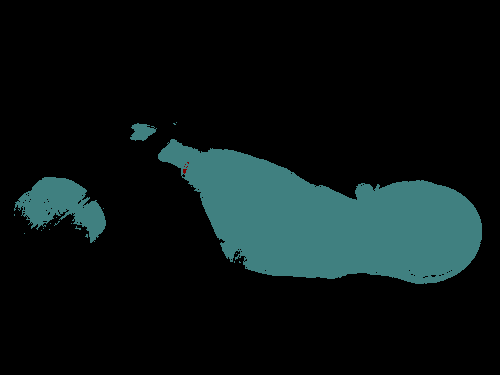}} & 
\multicolumn{1}{c}{\includegraphics[scale=0.1]{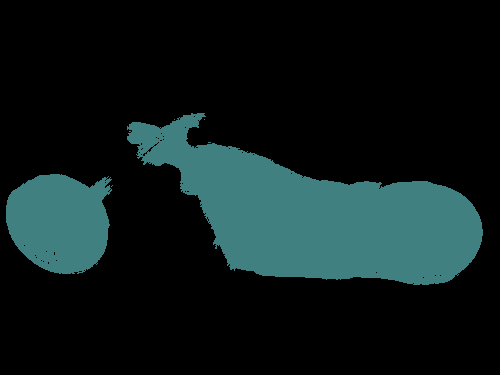}} &
\multicolumn{1}{c}{\includegraphics[scale=0.1]{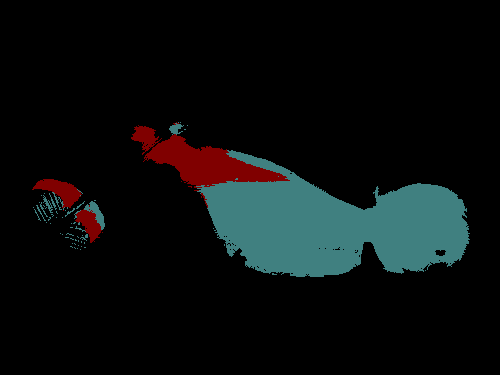}} & 

\multicolumn{1}{c}{\includegraphics[scale=0.1]{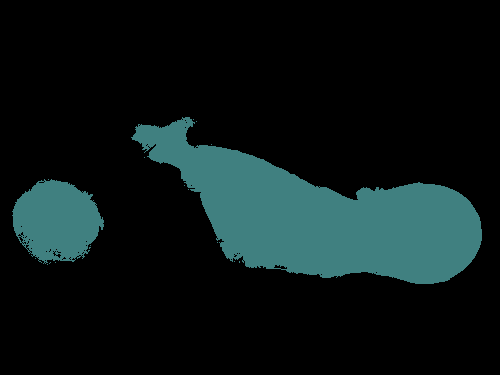}} & 
\multicolumn{1}{c}{\includegraphics[scale=0.1]{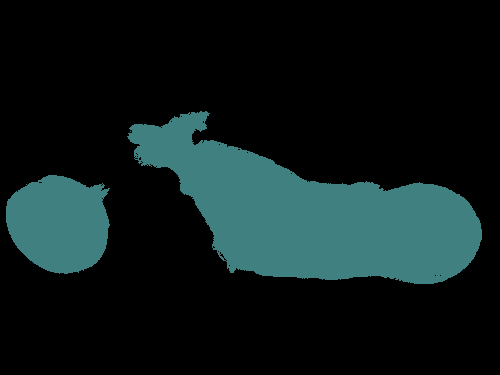}} &
\multicolumn{1}{c}{\includegraphics[scale=0.1]{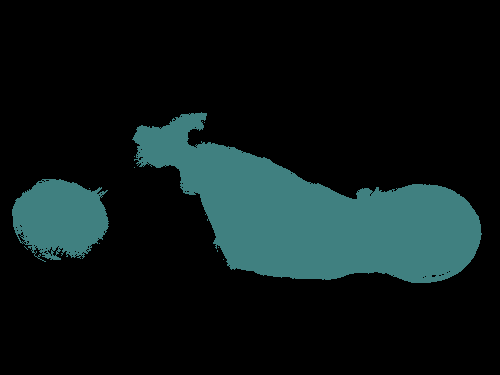}} \\ \hline
\multicolumn{1}{|c}{}& \multicolumn{1}{|c}{$S_1$} & \multicolumn{1}{c}{$S_2$} & \multicolumn{1}{c|}{$S_3$} &
\multicolumn{1}{|c}{$S_1$} & \multicolumn{1}{c}{$S_2$} & \multicolumn{1}{c|}{$S_3$} \\ \hline
\multicolumn{1}{|c}{Ground Truth}& \multicolumn{3}{|c|}{Adapt(\xmark)} & \multicolumn{3}{|c|}{Adapt (\cmark)} \\ \hline
\end{tabular}
\label{fig:adaptCRFComp}
\end{figure}

\begin{figure}[!ht]
\centering
\caption{Qualitative comparison of using \emph{adaptive} training. Images are taken from PASCAL VOC12 \emph{val} set and are \emph{not} post-processed with the CRF.}
\setlength\tabcolsep{1.0pt}
\begin{tabular}{|c|ccc|ccc|}
\multicolumn{1}{c}{\includegraphics[scale=0.1]{images/gt/2007_000346.png}} & 
\multicolumn{1}{c}{\includegraphics[scale=0.1]{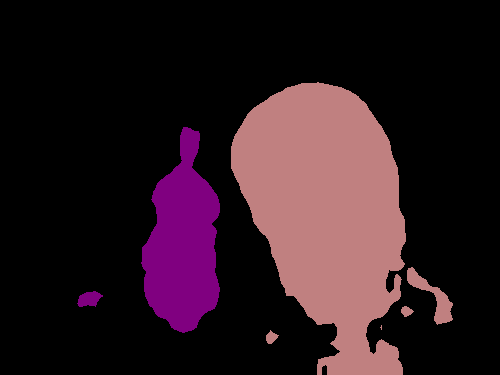}} & 
\multicolumn{1}{c}{\includegraphics[scale=0.1]{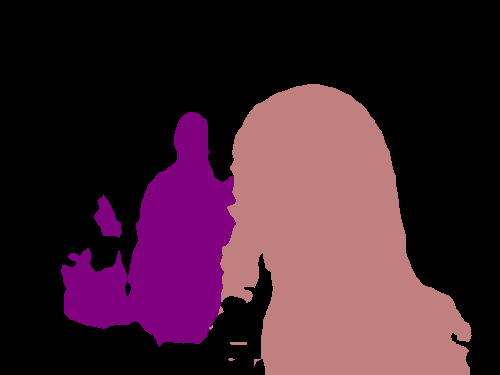}} &
\multicolumn{1}{c}{\includegraphics[scale=0.1]{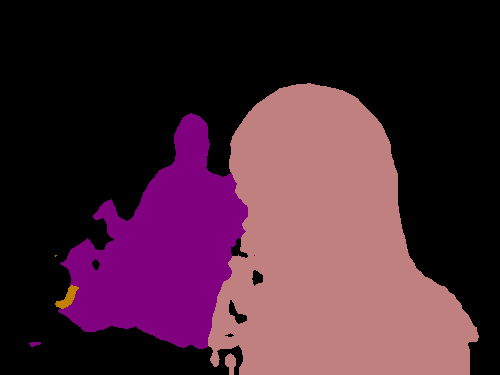}} & 

\multicolumn{1}{c}{\includegraphics[scale=0.1]{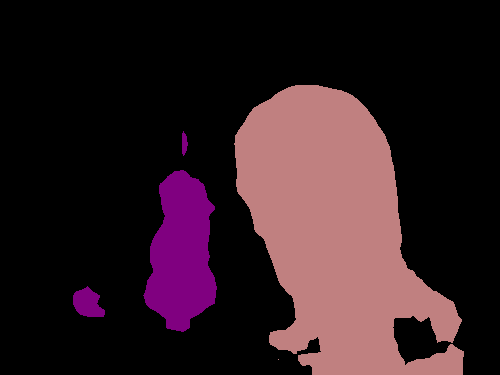}} & 
\multicolumn{1}{c}{\includegraphics[scale=0.1]{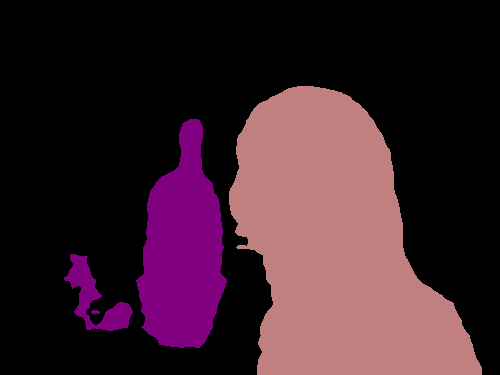}} &
\multicolumn{1}{c}{\includegraphics[scale=0.1]{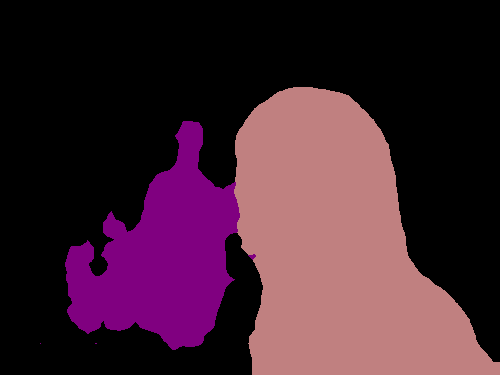}} \\

\multicolumn{1}{c}{\includegraphics[scale=0.1]{images/gt/2007_001288.png}} &
\multicolumn{1}{c}{\includegraphics[scale=0.1]{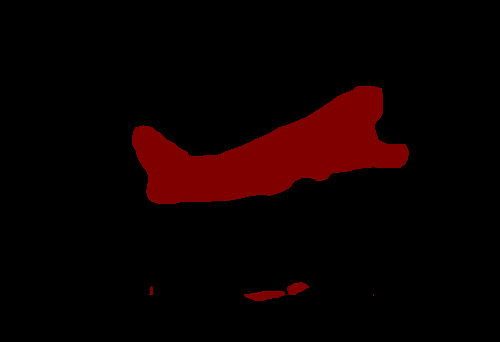}} & 
\multicolumn{1}{c}{\includegraphics[scale=0.1]{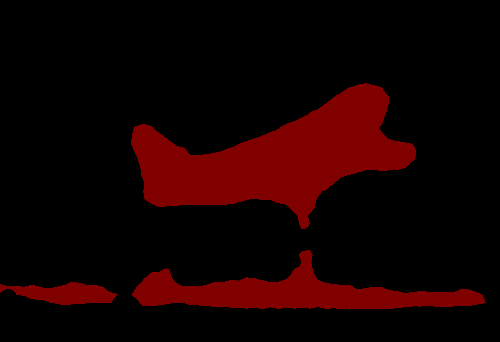}} &
\multicolumn{1}{c}{\includegraphics[scale=0.1]{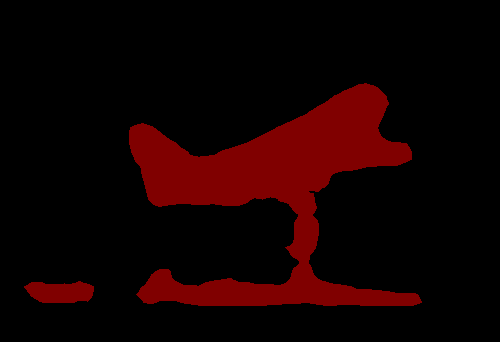}} & 

\multicolumn{1}{c}{\includegraphics[scale=0.1]{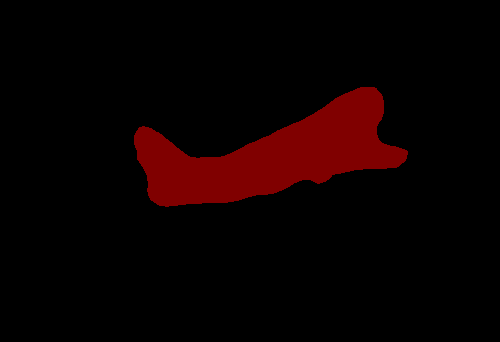}} & 
\multicolumn{1}{c}{\includegraphics[scale=0.1]{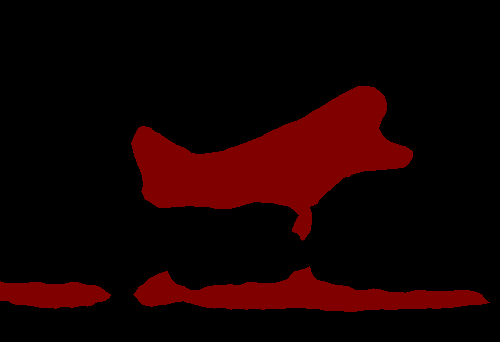}} &
\multicolumn{1}{c}{\includegraphics[scale=0.1]{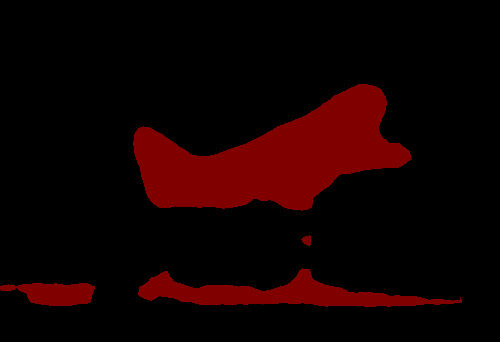}} \\

\multicolumn{1}{c}{\includegraphics[scale=0.1]{images/gt/2007_002823.png}} &
\multicolumn{1}{c}{\includegraphics[scale=0.1]{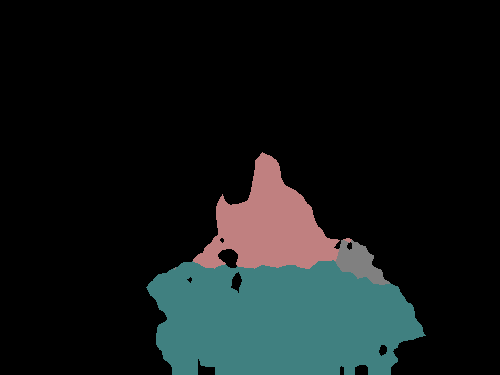}} & 
\multicolumn{1}{c}{\includegraphics[scale=0.1]{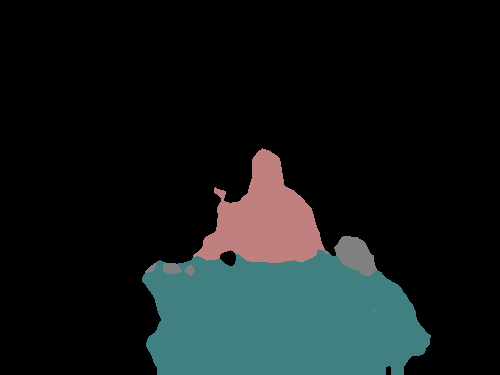}} &
\multicolumn{1}{c}{\includegraphics[scale=0.1]{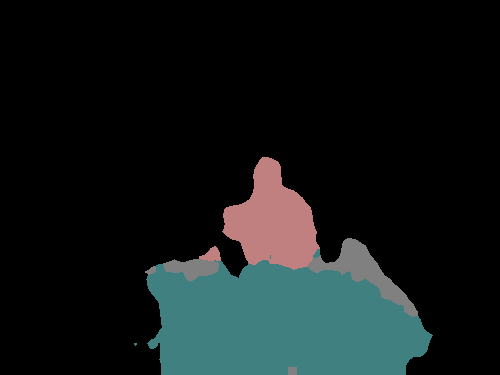}} & 

\multicolumn{1}{c}{\includegraphics[scale=0.1]{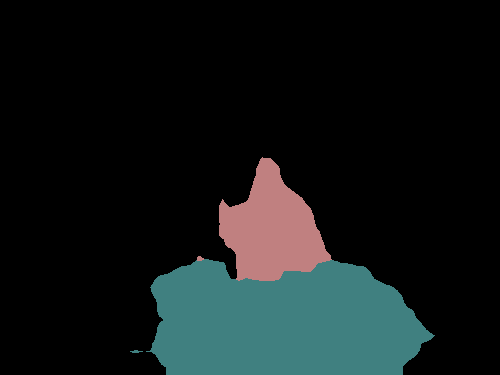}} & 
\multicolumn{1}{c}{\includegraphics[scale=0.1]{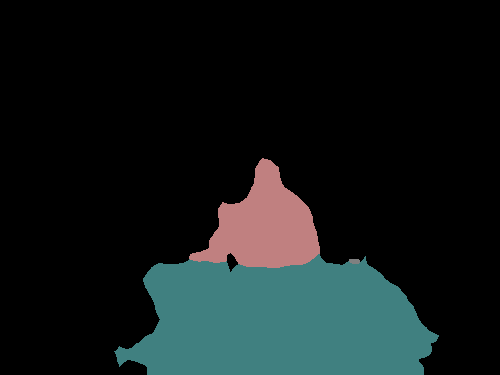}} &
\multicolumn{1}{c}{\includegraphics[scale=0.1]{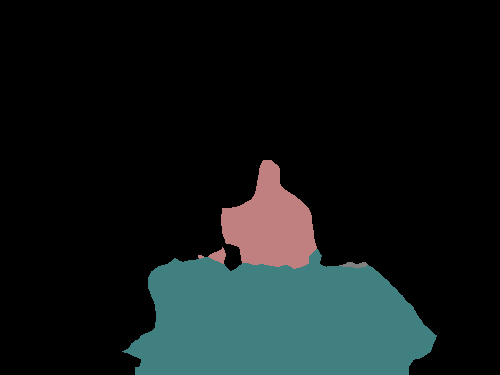}} \\

\multicolumn{1}{c}{\includegraphics[scale=0.1]{images/gt/2008_005637.png}} &
\multicolumn{1}{c}{\includegraphics[scale=0.1]{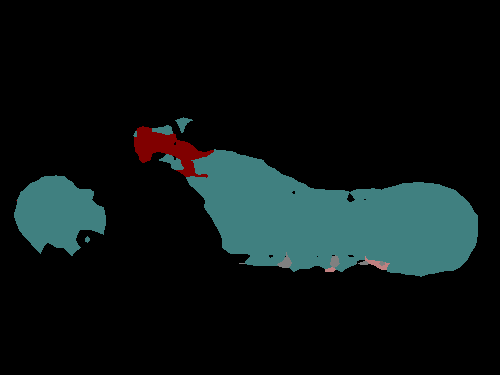}} & 
\multicolumn{1}{c}{\includegraphics[scale=0.1]{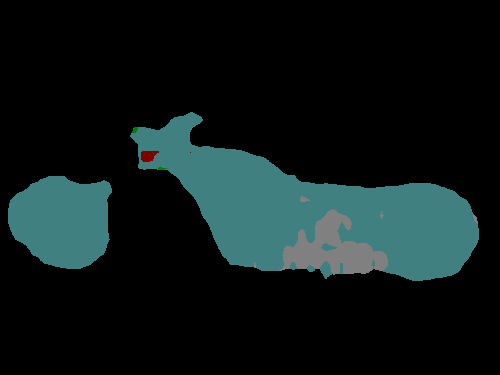}} &
\multicolumn{1}{c}{\includegraphics[scale=0.1]{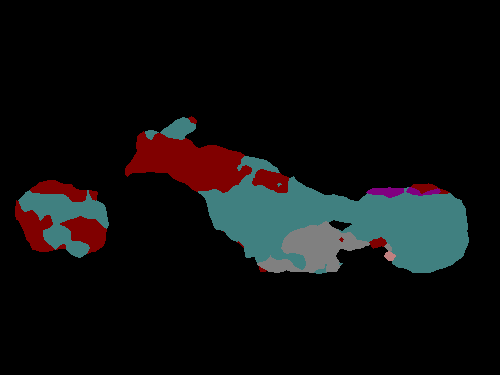}} & 

\multicolumn{1}{c}{\includegraphics[scale=0.1]{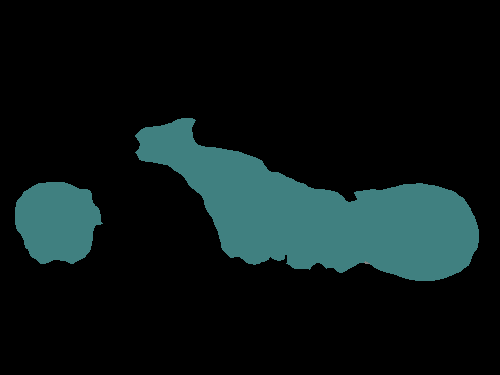}} & 
\multicolumn{1}{c}{\includegraphics[scale=0.1]{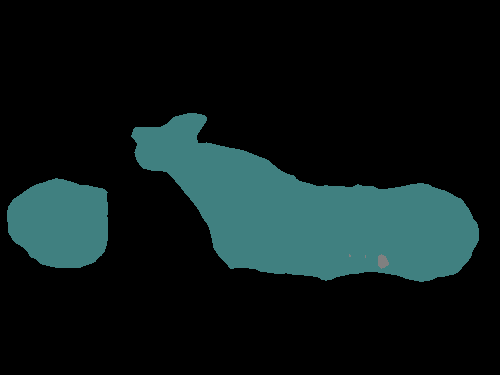}} &
\multicolumn{1}{c}{\includegraphics[scale=0.1]{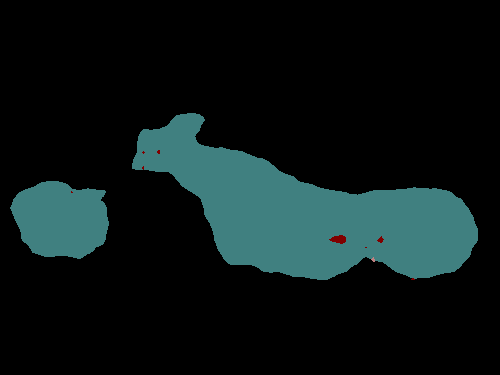}} \\ \hline
\multicolumn{1}{|c}{}& \multicolumn{1}{|c}{$S_1$} & \multicolumn{1}{c}{$S_2$} & \multicolumn{1}{c|}{$S_3$} &
\multicolumn{1}{|c}{$S_1$} & \multicolumn{1}{c}{$S_2$} & \multicolumn{1}{c|}{$S_3$} \\ \hline
\multicolumn{1}{|c}{Ground Truth}& \multicolumn{3}{|c|}{Adapt(\xmark)} & \multicolumn{3}{|c|}{Adapt (\cmark)} \\ \hline
\end{tabular}
\label{fig:adaptNoCRFComp}
\end{figure}

In Figures (\ref{fig:adaptCRFComp},~\ref{fig:adaptNoCRFComp}), we qualitatively compare the effects of adapting the ground-truth during the training. Recall that in the case of adaptive training we update the localization cues by taking the $\argmax$ over the categories present in the image on the segmentation volume (referred to as $f^m(I_i; \theta)$ in the main paper). As can be seen from the figures that enforcing the constraint of image-level labels at the adapt step allows us to remove many false-positive activations. For example, see how in the first two rows of the figures the background pixels erroneously assigned to foreground (person and plane, respectively) are corrected after the adapt step. Similarly, in the next two rows, extra classes are removed by the adaptive training. This suggests that using the output of the network constrained by the image labels at the adapt step produces more refined cues for training.

\begin{table}[htb]
\centering
\caption{Comparison of segmentation accuracies by our method (DCSP) for object categories using different \emph{Hierarchical Saliency} steps on PASCAL VOC 2012 \emph{val} set. $S_1$, $S_2$ and 
$S_3$ are the saliency maps of the original image, image after first erasing and image after second erasing, respectively.}
\label{tab:catgValMIOU}
\resizebox{13cm}{!}{%
\begin{tabular}{|c|cc|cc|cc|}
\hline
Category               & \multicolumn{2}{c|}{\begin{tabular}[c]{@{}c@{}}S1\\ IoU(\%)\end{tabular}} & \multicolumn{2}{c|}{\begin{tabular}[c]{@{}c@{}}S2\\ IoU(\%)\end{tabular}} & \multicolumn{2}{c|}{\begin{tabular}[c]{@{}c@{}}S3\\ IoU(\%)\end{tabular}} \\ \hline
\multicolumn{1}{|l|}{} & \multicolumn{1}{l|}{w/o CRF}           & \multicolumn{1}{l|}{w CRF}    & \multicolumn{1}{l|}{w/o CRF}            & \multicolumn{1}{l|}{w CRF}   & \multicolumn{1}{l|}{w/o CRF}            & \multicolumn{1}{l|}{w CRF}   \\ \hline
bcgd                   & 87.6                                   & 87.8                        & 88.5                                 & 89                       & 88.3                                 & 88.9                       \\
aeroplane              & 72                                    & 74.3                        & 73.2                                 & 76.3                       & 74.9                                 & 77.65                       \\
bicycle                & 29.3                                  & 28.2                        & 31.9                                 & 32.5                       & 31                                 & 31.3                       \\
bird                   & 75.4                                   & 77.7                        & 71.3                                 & 74.5                       & 69.3                                 & 73.2                       \\
boat                   & 58.4                                   & 59                          & 59.1                                 & 60.8                       & 58.3                                 & 59.8                       \\
bottle                 & 63.7                                   & 64.1                        & 67.8                                 & 69.8                       & 69.4                                 & 71.0                       \\
bus                    & 61.7                                   & 62.2                        & 74.3                                 & 74.9                       & 77.6                                 & 79.2                       \\
car                    & 68.5                                   & 69.3                        & 72.9                                 & 74.1                       & 72.3                                 & 74.5                       \\
cat                    & 80.4                                   & 83                          & 79.7                                 & 82.5                       & 77.9                                 & 80                       \\
chair                  & 11.6                                   & 10.7                        & 14.4                                 & 13.2                       & 16.4                                   & 15.1                       \\
cow                    & 69.1                                   & 70.6                        & 73.3                                 & 75.3                       & 71.4                                 & 73.3                       \\
diningtable            & 3.6                                    & 3                           & 6.8                                  & 6.3                       & 12                                 & 10.2                       \\
dog                    & 74.8                                   & 76.8                        & 74.5                                 & 76.9                       & 74.1                                 & 76.1                       \\
horse                  & 62.9                                   & 64.3                        & 71.4                                 & 74.7                       & 69.3                                 & 72.21                       \\
motorbike              & 64.5                                   & 64.9                        & 69.1                                 & 70.1                         & 68                                 & 69.1                      \\
person                 & 66.6                                   & 67.6                        & 70.1                                 & 71.4                       & 70.5                                 & 72.1                       \\
pottedplant            & 34                                    & 33.8                        & 39.7                                 & 40                       & 39.2                                 & 39.9                       \\
sheep                  & 63.4                                   & 64.2                        & 70.4                                 & 73                       & 70.7                                 & 73.9                       \\
sofa                   & 12.6                                   & 12.3                        & 17.1                                 & 17.1                       & 15.8                                   & 14.6                       \\
train                  & 58.7                                   & 57.5                        & 71.4                                 & 72.3                       & 69.8                                 & 70.3                       \\
tvmonitor              & 51.8                                   & 51.4                        & 52.3                                 & 53.4                       & 52.6                                 & 53.1                       \\ \hline
Average                & \multicolumn{1}{c|}{\textbf{55.7}}     & \textbf{56.3}                 & \multicolumn{1}{c|}{\textbf{59.5}}   & \textbf{60.8}              & \multicolumn{1}{c|}{\textbf{59.5}}   & \textbf{60.8}              \\ \hline
\end{tabular}%
}
\end{table}

\begin{table}[htb]
\centering
\caption{Comparison of segmentation accuracies by our method (DCSP) for object categories using different \emph{Hierarchical Saliency} steps on PASCAL VOC 2012 \emph{test} set. $S_1$, $S_2$ and 
$S_3$ are the saliency maps of the original image, image after first erasing and image after second erasing, respectively.}
\label{tab:catgTestMIOU}
\resizebox{13cm}{!}{%
\begin{tabular}{|c|cc|cc|cc|}
\hline
Category               & \multicolumn{2}{c|}{\begin{tabular}[c]{@{}c@{}}S1\\ IoU(\%)\end{tabular}} & \multicolumn{2}{c|}{\begin{tabular}[c]{@{}c@{}}S2\\ IoU(\%)\end{tabular}} & \multicolumn{2}{c|}{\begin{tabular}[c]{@{}c@{}}S3\\ IoU(\%)\end{tabular}} \\ \hline
\multicolumn{1}{|l|}{} & \multicolumn{1}{l|}{w/o CRF}           & \multicolumn{1}{l|}{w CRF}    & \multicolumn{1}{l|}{w/o CRF}            & \multicolumn{1}{l|}{w CRF}   & \multicolumn{1}{l|}{w/o CRF}            & \multicolumn{1}{l|}{w CRF}   \\ \hline
bcgd                   & 88.3                                   & 88.5                        & 88.9\                                 & 89.3                       & 88.8                                 & 89.3                       \\
aeroplane              & 72.4                                   & 73.7                        & 73.1\                                 & 75.6                       & 76.7                                 & 79.4                       \\
bicycle                & 29.6                                   & 29.2                        & 30.7\                                 & 31.6                       & 31.4                                 & 32.5                       \\
bird                   & 73.1                                    & 75.7                        & 68.6\                                 & 71.3                       & 69.3                                 & 72.9                       \\
boat                   & 49.0                                   & 49.5                        & 51.5\                                 & 53.2                       & 49.7                                 & 51.7                       \\
bottle                 & 63                                   & 63.5                        & 66.6\                                 & 68.2                       & 64.4                                 & 66.4                       \\
bus                    & 61.6                                   & 61.7                        & 74.2                                 & 75.3                       & 76.7                                 & 77.2                       \\
car                    & 74.9                                   & 75.8                        & 75.9                                 & 76.9                       & 75.9                                 & 77.3                       \\
cat                    & 77.1                                   & 79.1                        & 80                                   & 82.5                       & 78.4                                 & 81.5                       \\
chair                  & 14.9                                   & 14.9                        & 21.2                                 & 20.9                       & 21                                   & 20.8                       \\
cow                    & 71.9                                   & 74.8                        & 73.7                                 & 75.6                       & 72.9                                 & 75.6                       \\
diningtable            & 4.7                                   & 4.01                        & 13.5                                 & 12.2                       & 14.8                                 & 12.9                       \\
dog                    & 76.9                                   & 79.0                        & 75.1                                 & 77.8                       & 75.8                                 & 79.3                       \\
horse                  & 71.8                                   & 73.8                        & 73.7                                 & 76.2                       & 71.4                                 & 74.5                       \\
motorbike              & 68.2                                   & 67.8                        & 75.5                                 & 77                         & 75.2                                 & 76.9                       \\
person                 & 68.1                                   & 69.3                        & 69.8                                 & 71.5                       & 70.2                                 & 71.8                       \\
pottedplant            & 32.2                                   & 32.7                        & 39.1                                 & 38.9                       & 39.9                                 & 39.3                       \\
sheep                  & 73.4                                   & 75.9                        & 78.6                                 & 82.2                       & 77.8                                 & 81.7                       \\
sofa                   & 13.4                                   & 13.1                        & 25.3                                 & 25.1                       & 25                                   & 24.3                       \\
train                  & 56.9                                   & 55.7                        & 63.6                                 & 63.8                       & 63.9                                 & 63.9                       \\
tvmonitor              & 46.9                                   & 46.7                        & 47.2                                 & 48.7                       & 48.1                                 & 49.8                       \\ \hline
Average                & \multicolumn{1}{c|}{\textbf{56.6}}     & \textbf{57.3}                 & \multicolumn{1}{c|}{\textbf{60.2}}   & \textbf{61.6}              & \multicolumn{1}{c|}{\textbf{60.3}}   & \textbf{61.9}              \\ \hline
\end{tabular}%
}
\end{table}

\clearpage

\bibliography{arslanBibliography.bib}
\end{document}